\documentclass[final,3p,times]{elsarticle}

\usepackage{amssymb}
\usepackage{amsmath}
\usepackage{graphicx}
\usepackage{multirow}
\usepackage{booktabs}
\usepackage{algpseudocode}
\usepackage{algorithm}
\usepackage{url}

\journal{}

\begin{document}

\begin{frontmatter}

\title{PQFA: Parallel Quantum Feature Augmentation of Fused Representations for Multimodal Classification}

\author[1,2]{Mingzhu Wang} 
\author[1,3]{Yun Shang\corref{cor1}}
\ead{shangyun@amss.ac.cn}
\affiliation[1]{organization={Academy of Mathematics and Systems Science, Chinese Academy of Sciences},
            city={Beijing},
            postcode={100190}, 
            country={China}}
\affiliation[2]{organization={School of Mathematical Sciences, University of Chinese Academy of Sciences},
            city={Beijing},
            postcode={100049}, 
            country={China}}
\affiliation[3]{organization={State Key Laboratory of Mathematical Sciences, Academy of Mathematics and Systems Science, Chinese Academy of Sciences},
            city={Beijing},
            postcode={100190}, 
            country={China}}
\cortext[cor1]{Corresponding author}
\begin{abstract}
Most multimodal learning methods improve how heterogeneous representations are aligned and fused, while post-fusion enhancement remains less explored. We propose Parallel Quantum Feature Augmentation (PQFA), a hybrid quantum--classical framework that applies multiple shallow variational quantum circuits to fused multimodal features. Text and image representations extracted by frozen RoBERTa and ViT encoders are processed through bidirectional cross-attention, attentive pooling, and adaptive gated fusion. The fused feature is then amplitude-encoded into parallel quantum circuits, whose measurement readouts are concatenated with the classical representation for prediction. We evaluate PQFA on MM-IMDb and N24News through controlled comparisons using the same encoders, fusion backbone, data splits, projection dimension, and augmentation output width. PQFA consistently outperforms both the fusion backbone without quantum augmentation and a width-matched MLP augmentation baseline, while using approximately 2.2K augmentation parameters compared with 24.0K for the MLP branch. Missing-modality experiments further show improved robustness when textual or visual inputs are incomplete, with particularly clear gains when the more informative textual modality is severely degraded. Controlled ablations and feature-space analyses indicate that the improvement cannot be reproduced by random feature mappings, increased classical width, or untrained quantum transformations. Quantum-state diagnostics additionally show stable predictive performance across the tested simulated noise levels and distinct branch-specific transformations of the encoded states. These results establish PQFA as an effective and parameter-efficient strategy for post-fusion augmentation in hybrid quantum--classical multimodal learning.
\end{abstract}


\end{frontmatter}

\section{Introduction}

Multimodal classification requires a model to integrate heterogeneous information from different sources, such as text and images. 
Modern multimodal models usually rely on pretrained encoders and intermediate fusion modules to align modality-specific representations before prediction~\cite{baltruvsaitis2018multimodal,xu2023multimodal}. 
Although these classical fusion architectures are effective, the fused representation may still benefit from lightweight post-fusion transformations that enrich features useful for prediction without substantially increasing the depth or size of the whole model. Compared with the extensive literature on improving the fusion stage itself, post-fusion augmentation modules that operate on an already fused representation remain relatively underexplored. Moreover, the contribution of such a module must be carefully separated from that of the underlying fusion backbone.

Parameterized quantum circuits provide a possible mechanism for such post-fusion feature transformation. 
A variational quantum circuit can encode a classical vector into a quantum state, apply a trainable unitary transformation, and return measurement expectations as nonlinear readout features~\cite{biamonte2017quantum,schuld2019quantum,havlivcek2019supervised}. 
However, current noisy intermediate-scale (NISQ) quantum settings impose practical constraints on qubit number, circuit depth, finite-shot noise, and trainability~\cite{preskill2018quantum,bharti2022noisy,mcclean2018barren}. 
These constraints make it impractical to replace mature deep multimodal architectures with large quantum models. 
For near-term settings, a more appropriate use of quantum circuits is to treat them as shallow auxiliary modules within an otherwise classical pipeline.

Existing quantum or quantum-inspired multimodal studies have explored quantum fusion modules, quantum-enhanced sentiment analysis, and hybrid neural architectures for multimodal tasks~\cite{qu2024qmfnd,qu2023qnmf,zheng2024quantum,li2025qmlsc}. 
Nevertheless, existing studies do not always clearly isolate the architectural role and contribution of the quantum component. 
Many designs process unimodal features separately or insert quantum modules as sequential blocks, making it unclear whether the quantum component improves an already integrated cross-modal representation. 
Moreover, reported gains are often entangled with increased feature dimensionality, additional trainable parameters, or changes in the classical fusion backbone.

This distinction is important because a performance gain in a hybrid quantum--classical multimodal model can have several different causes. 
It may result from a better fusion strategy, a stronger pretrained encoder, a larger final representation, or simply a larger number of trainable parameters. 
Without a backbone that removes the quantum circuits and a width-matched classical augmentation baseline, it is difficult to attribute the improvement specifically to the quantum readout branch. 
Therefore, a meaningful evaluation of quantum-enhanced multimodal fusion should control the classical fusion pathway and compare the quantum module with a classical module of the same output width.
Beyond aggregate performance, it is also useful to inspect the geometry of the generated augmentation features. Comparing PCA spectra, effective dimensions, and structurally modified augmentation branches can help distinguish a task-aligned feature transformation from generic width expansion or random nonlinear feature generation.

To address this issue, we propose Parallel Quantum Feature Augmentation (PQFA). 
The central idea is not to replace the classical multimodal fusion module, but to augment the fused representation after cross-modal interaction has already been performed. 
Multimodal inputs are first encoded by frozen pretrained encoders, refined by bidirectional cross-attention, summarized by attentive pooling, and integrated through adaptive gated fusion. 
The resulting fused feature is then amplitude-encoded into several shallow variational quantum circuits in parallel. 
Their measurement readouts are concatenated with the classical fused feature for final prediction.

The main contributions of this work are summarized as follows:
\begin{itemize}
\item We introduce PQFA, a post-fusion quantum augmentation framework that applies shallow parallel variational quantum circuits to an integrated multimodal representation rather than to raw or independently processed unimodal features.

\item We develop a controlled hybrid architecture in which the pretrained encoders and classical fusion pathway are preserved across PQFA, the no-quantum backbone, and the classical augmentation baselines. This design strengthens the attribution of performance differences to the augmentation module.

\item We demonstrate on MM-IMDb and N24News that PQFA improves multi-label and single-label classification while using substantially fewer augmentation parameters than a width-matched MLP branch. Additional controls distinguish the learned quantum transformation from generic feature expansion, increased classical capacity, and random quantum mappings.

\item We evaluate PQFA beyond aggregate predictive metrics through missing-modality experiments, paired error-transition analysis, feature-space diagnostics, simulated-noise evaluation, entanglement analysis, and gate-weight inspection. These analyses characterize its robustness, decision-level behavior, and representation properties.

\end{itemize}

The remainder of this paper is organized as follows.
Section~\ref{sec:related} reviews classical multimodal fusion, hybrid quantum--classical learning, and quantum-enhanced multimodal models.
Section~\ref{sec:method} presents the PQFA architecture.
Section~\ref{sec:experiments} describes the experimental settings and reports the main results and diagnostic analyses.
Section~\ref{sec:discussion} discusses the implications and future research directions.
Section~\ref{sec:conclusion} concludes the paper.

\section{Related Work}
\label{sec:related}

This section reviews three lines of research most relevant to the present study: classical multimodal fusion, quantum machine learning and hybrid quantum--classical models, and quantum or quantum-inspired multimodal fusion.

\subsection{Classical Multimodal Fusion}

Multimodal fusion aims to integrate heterogeneous information sources into a unified representation for prediction and decision making.
Existing strategies are commonly categorized as early, intermediate, and late fusion according to whether modalities are combined at the input, representation, or decision level~\cite{gao2020survey,jiao2024comprehensive}.
Early fusion is simple but can be sensitive to feature-scale mismatch and modality heterogeneity, whereas late fusion preserves modality-specific pathways but may underuse fine-grained cross-modal interactions~\cite{gao2020survey,jiao2024comprehensive}.
Intermediate fusion has therefore become a widely adopted paradigm in deep multimodal learning because it permits explicit alignment and interaction between modality-specific representations~\cite{jiao2024comprehensive,xu2023multimodal}.

A major line of work focuses on increasing the expressiveness of cross-modal interactions while controlling computational cost.
Low-rank and factorized methods approximate high-order fusion tensors through modality-specific factors~\cite{liu2018efficient}.
In vision--language learning, Multimodal Factorized Bilinear Pooling and Multimodal Tucker Fusion further model multiplicative image--text interactions using compact bilinear or tensor decompositions~\cite{yu2017multi,ben2017mutan}.
Other approaches learn shared and modality-specific factors, maximize cross-modal correlations, or iteratively refine unimodal representations before prediction~\cite{sun2020learning,guo2021unimodal}.
Transformer-based models instead use self-attention, cross-attention, and attention bottlenecks to exchange information among tokens, image patches, and modality-specific latent features~\cite{vaswani2017attention,nagrani2021attention,xu2023multimodal}.
Dynamic gating and coupled state-space models further adapt the contribution or interaction pattern of each modality according to the input~\cite{xue2023dynamic,li2024coupled}.

These developments are directly relevant to image--text classification, where cross-attention can establish semantic correspondences and adaptive gating can regulate the relative contribution of textual and visual representations~\cite{xu2023multimodal,xue2023dynamic}.
Robustness studies also show that multimodal Transformers can be sensitive to missing inputs and that the effect of a fusion strategy may depend on the dataset and missing-modality pattern~\cite{ma2022multimodal}.
Most of this literature, however, improves the encoders, interaction mechanism, or fusion operator itself.
Comparatively less attention has been paid to preserving a fixed, strong fusion backbone and asking whether an additional lightweight transformation can improve the already fused representation.
This distinction motivates the post-fusion setting considered in this work and supports comparisons against both the unaugmented fusion backbone and a width-matched classical augmentation module.

\subsection{Quantum Machine Learning and Hybrid Quantum--Classical Models}

Quantum machine learning studies how quantum states, measurements, and trainable quantum operations can be used for representation, learning, and prediction~\cite{biamonte2017quantum,schuld2019quantum}.
For near-term applications, parameterized quantum circuits (PQCs), also called variational quantum circuits, are commonly embedded in hybrid quantum--classical optimization loops~\cite{peruzzo2014variational,bharti2022noisy}.
A typical PQC-based learning model encodes classical data into a quantum state, applies a trainable unitary transformation, and converts measurement expectations into features or prediction outputs~\cite{schuld2019quantum,havlivcek2019supervised}.
This framework has motivated quantum kernels, variational classifiers, quantum neural networks, and other hybrid models for supervised learning and representation learning~\cite{havlivcek2019supervised,zaman2024comparative,basilewitsch2025quantum}.

The appeal of PQCs lies in their ability to implement structured nonlinear feature maps through data encoding, entangling operations, and observable measurements~\cite{schuld2019quantum,havlivcek2019supervised}.
Their practical use is nevertheless constrained by the cost of loading classical data, limited qubit counts and connectivity, finite circuit depth, sampling noise, and hardware errors~\cite{preskill2018quantum,bharti2022noisy}.
Optimization can also deteriorate when circuits are too deep, poorly initialized, or affected by noise, as reflected in barren-plateau and noise-induced trainability analyses~\cite{mcclean2018barren,wang2021noise,larocca2025barren}.
These limitations make the wholesale replacement of mature deep neural networks by large quantum models unrealistic in current settings.

Recent work therefore increasingly studies quantum components as modules within larger classical systems rather than as stand-alone replacements.
Such components have been used as compact feature transformations, classifier heads, local quantum filters, or adaptively structured circuit blocks~\cite{zhao2025hqcc,rizvi2025quantum,basilewitsch2025quantum}.
At the same time, controlled benchmarking has emphasized the need to compare quantum models with suitably tuned classical baselines and to separate architectural changes from the effect of the quantum component itself~\cite{schnabel2025quantum,basilewitsch2025quantum}.
This modular and controlled perspective is central to the present study.
PQFA retains pretrained classical encoders and a fixed cross-modal fusion pathway, while shallow parallel PQCs are evaluated specifically as post-fusion readout transformations.
The corresponding no-quantum and width-matched classical baselines are designed to distinguish quantum feature augmentation from generic increases in representation width or parameter count.

\subsection{Quantum and Quantum-Inspired Multimodal Fusion}

Quantum-enhanced multimodal learning lies at the intersection of multimodal representation learning and hybrid QML and remains substantially less developed than either area separately.
Existing studies have explored several placements and physical interpretations of quantum components.
QMFND combines textual and visual information for fake-news detection using a quantum multimodal fusion architecture~\cite{qu2024qmfnd}, while QNMF studies quantum-neural multimodal fusion for intelligent diagnosis~\cite{qu2023qnmf}.
Quantum circuit models have also been proposed for multimodal sentiment classification, and recent work has investigated explicit quantum fusion layers intended to represent higher-order interactions across modalities~\cite{zheng2024quantum,li2025qmlsc,nguyen2025expressive}.
Related quantum-inspired approaches use density-matrix, open-system, or Lindblad-style dynamics to formulate multimodal interactions without necessarily implementing the complete model as an executable quantum circuit~\cite{yan2025quantum}.
Together, these studies indicate growing interest in using quantum or quantum-inspired structures to transform heterogeneous representations.

Despite this progress, the architectural role of the quantum component is not yet evaluated consistently.
Across existing models, quantum circuits may process unimodal features, participate directly in fusion, or appear as sequential blocks after classical feature extraction~\cite{qu2023qnmf,qu2024qmfnd,zheng2024quantum,li2025qmlsc}.
These different placements make it difficult to determine whether a reported improvement originates from the quantum transformation, the fusion mechanism, the pretrained encoders, or an accompanying increase in feature dimension and trainable parameters.
The distinction between executable PQC-based models and quantum-inspired neural formulations also needs to be maintained because their computational assumptions and hardware implications differ~\cite{nguyen2025expressive,yan2025quantum}.

A further limitation is the scarcity of attribution-oriented evaluation.
Among the above studies, systematic comparisons using both an unchanged no-quantum fusion backbone and a classical augmentation module matched in output width remain uncommon.
Likewise, missing-modality robustness, feature-space behavior, circuit ablations, noise sensitivity, and sample-level decision changes are rarely examined together in quantum multimodal classification, despite their relevance to multimodal robustness and near-term QML~\cite{ma2022multimodal,wang2021noise,larocca2025barren}.
The present work addresses this gap by applying multiple shallow PQCs only after classical cross-modal fusion and by evaluating them against controlled no-quantum and width-matched classical alternatives.
This design also motivates the subsequent ablation, missing-modality, paired-transition, feature-space, and quantum-state analyses.

\section{Method}
\label{sec:method}

\begin{figure*}[t]
  \centering
  \includegraphics[width=\textwidth]{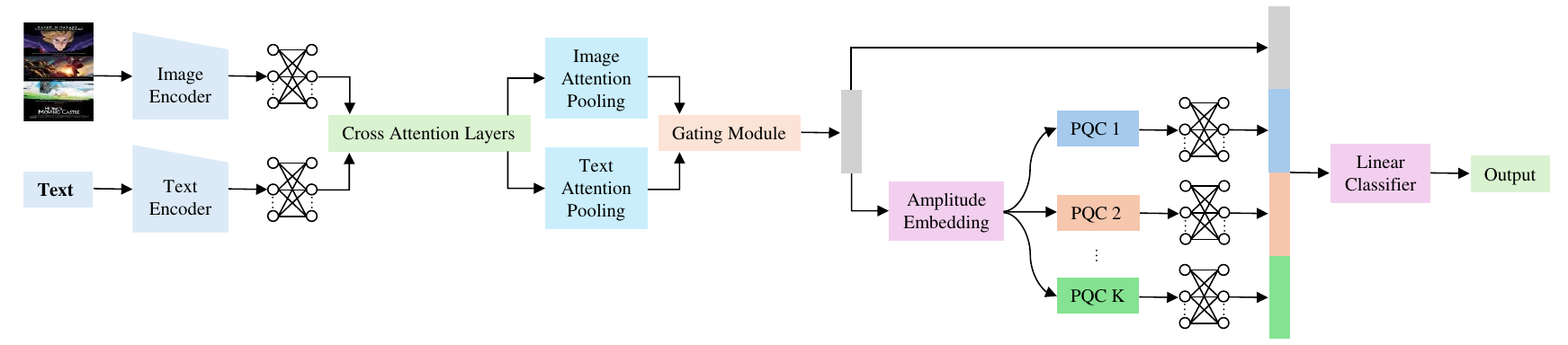}
  \caption{
  Overall architecture of the proposed Parallel Quantum Feature Augmentation (PQFA) framework. Text and image inputs are first encoded by frozen pretrained Transformer encoders. The extracted unimodal token features are projected into a shared latent space, refined by bidirectional cross-attention, summarized by attentive pooling, and fused through an adaptive gating module. The fused classical representation is then processed by two paths: a direct classical path and a parallel quantum augmentation path based on amplitude encoding and multiple shallow parameterized quantum circuits. The resulting quantum readouts are concatenated with the classical fused feature for final classification.
  }
  \label{fig:pqfa_framework}
\end{figure*}

\subsection{Problem Formulation}

We formulate PQFA for paired multimodal classification. 
Fig.~\ref{fig:pqfa_framework} illustrates how PQFA constructs fused multimodal representations and augments them with parallel quantum readout branches.
Let
\begin{equation}
\label{eq:dataset}
\mathcal{D}
=
\{(X_n^{(1)},X_n^{(2)},y_n)\}_{n=1}^{N}
\end{equation}
denote a multimodal dataset, where $X_n^{(1)}$ and $X_n^{(2)}$ are paired inputs and $y_n$ is the target. 
The target can take either a multi-label or a single-label form. 
For multi-label classification, $y_n\in\{0,1\}^{C}$ is a binary label vector over $C$ classes. 
For single-label classification, $y_n\in\{1,\ldots,C\}$ is a categorical class index.

The model first constructs a fused multimodal representation
\begin{equation}
\label{eq:fused_representation_function}
\mathbf{z}_{C,n}
=
F_{\mathrm{fuse}}
\left(
X_n^{(1)},X_n^{(2)};
\boldsymbol{\Theta}_{\mathrm{fuse}}
\right),
\end{equation}
and then augments it with parallel quantum readout features:
\begin{equation}
\label{eq:quantum_aug_function}
\mathbf{q}_{n}
=
F_{q}
\left(
\mathbf{z}_{C,n};
\boldsymbol{\Theta}_{q}
\right).
\end{equation}
The final representation is
\begin{equation}
\label{eq:method_final_representation}
\mathbf{h}_{n}
=
[\mathbf{z}_{C,n};\mathbf{q}_{n}],
\end{equation}
which is passed to a task-specific prediction head. 
Thus, the quantum circuits are applied to fused representations rather than to raw or weakly aligned unimodal features.

\subsection{Frozen Unimodal Encoding and Feature Alignment}

In this work, \(X^{(1)}\) corresponds to text and \(X^{(2)}\) corresponds to image. Given a tokenized text input \(T\), a pretrained language encoder \(E_T\) produces contextual token features:
\begin{equation}
\label{eq:text_features}
\mathbf{X}_T = E_T(T) \in \mathbb{R}^{L_T\times d_T},
\end{equation}
where \(L_T\) is the number of text tokens and \(d_T\) is the hidden dimension of the text encoder. For the image input \(I\), a pretrained vision encoder \(E_I\) extracts patch-level visual features:
\begin{equation}
\label{eq:image_features}
\mathbf{X}_I = E_I(I) \in \mathbb{R}^{L_I\times d_I},
\end{equation}
where \(L_I\) denotes the number of visual tokens or patches and \(d_I\) is the hidden dimension of the vision encoder.

To align the two modalities into a common latent space, we apply learnable linear projections followed by layer normalization:
\begin{equation}
\label{eq:text_projection}
\mathbf{H}_T^{(0)}
=
\mathrm{LN}
\left(
\mathbf{X}_T\mathbf{W}_T+\mathbf{1}\mathbf{b}_T^{\top}
\right),
\qquad
\mathbf{W}_T\in\mathbb{R}^{d_T\times d},
\end{equation}
\begin{equation}
\label{eq:image_projection}
\mathbf{H}_I^{(0)}
=
\mathrm{LN}
\left(
\mathbf{X}_I\mathbf{W}_I+\mathbf{1}\mathbf{b}_I^{\top}
\right),
\qquad
\mathbf{W}_I\in\mathbb{R}^{d_I\times d}.
\end{equation}
Here, \(d\) is the shared embedding dimension, \(\mathrm{LN}(\cdot)\) denotes layer normalization, and \(\mathbf{1}\) is a vector of ones used to broadcast the bias terms across tokens.

\subsection{Bidirectional Cross-Modal Refinement}

The aligned unimodal sequences are refined by stacked bidirectional cross-attention layers. This module follows the standard Transformer attention formulation \cite{vaswani2017attention}, but uses it in a cross-modal manner: the query comes from one modality, while the keys and values come from the other modality.

At layer \(\ell\), the text stream queries the visual stream:
\begin{equation}
\label{eq:text_cross_attention}
\widetilde{\mathbf{H}}_T^{(\ell)}
=
\mathrm{LN}
\left(
\mathbf{H}_T^{(\ell)}
+
\mathrm{MHA}_T^{(\ell)}
\left(
\mathbf{H}_T^{(\ell)},
\mathbf{H}_I^{(\ell)},
\mathbf{H}_I^{(\ell)}
\right)
\right),
\end{equation}
\begin{equation}
\label{eq:text_ffn_update}
\mathbf{H}_T^{(\ell+1)}
=
\mathrm{LN}
\left(
\widetilde{\mathbf{H}}_T^{(\ell)}
+
\mathrm{FFN}_T^{(\ell)}
\left(
\widetilde{\mathbf{H}}_T^{(\ell)}
\right)
\right).
\end{equation}
The visual stream is updated by attending to the refined text stream:
\begin{equation}
\label{eq:image_cross_attention}
\widetilde{\mathbf{H}}_I^{(\ell)}
=
\mathrm{LN}
\left(
\mathbf{H}_I^{(\ell)}
+
\mathrm{MHA}_I^{(\ell)}
\left(
\mathbf{H}_I^{(\ell)},
\mathbf{H}_T^{(\ell+1)},
\mathbf{H}_T^{(\ell+1)}
\right)
\right),
\end{equation}
\begin{equation}
\label{eq:image_ffn_update}
\mathbf{H}_I^{(\ell+1)}
=
\mathrm{LN}
\left(
\widetilde{\mathbf{H}}_I^{(\ell)}
+
\mathrm{FFN}_I^{(\ell)}
\left(
\widetilde{\mathbf{H}}_I^{(\ell)}
\right)
\right).
\end{equation}

For an input sequence \(\mathbf{X}\in\mathbb{R}^{L\times d}\), the position-wise feed-forward network is defined as
\begin{equation}
\label{eq:ffn_definition}
\mathrm{FFN}(\mathbf{X})
=
\phi(\mathbf{X}\mathbf{W}_{1}+\mathbf{b}_{1})\mathbf{W}_{2}+\mathbf{b}_{2},
\end{equation}
where \(\phi(\cdot)\) is a nonlinear activation function. Unlike attention, which mixes information across tokens or modalities, the FFN refines each token representation along the feature dimension.

For a query sequence \(\mathbf{Q}\), key sequence \(\mathbf{K}\), and value sequence \(\mathbf{V}\), the \(r\)-th attention head is computed as
\begin{equation}
\label{eq:attention_head}
\mathrm{head}_r
=
\mathrm{softmax}
\left(
\frac{
(\mathbf{Q}\mathbf{W}_{r}^{Q})
(\mathbf{K}\mathbf{W}_{r}^{K})^{\top}
}{
\sqrt{d_h}
}
\right)
\mathbf{V}\mathbf{W}_{r}^{V},
\end{equation}
and the multi-head attention output is
\begin{equation}
\label{eq:mha_definition}
\mathrm{MHA}(\mathbf{Q},\mathbf{K},\mathbf{V})
=
\mathrm{Concat}
\left(
\mathrm{head}_1,\ldots,\mathrm{head}_H
\right)
\mathbf{W}^{O},
\end{equation}
where \(H\) is the number of attention heads and \(d_h=d/H\) is the dimension of each head. Through this bidirectional refinement process, token-level textual semantics and patch-level visual cues are repeatedly aligned before global fusion.

\subsection{Attentive Global Pooling}

After \(S\) cross-modal refinement layers, we obtain the final token sequences:
\begin{equation}
\label{eq:final_token_sequences}
\mathbf{H}_T=\mathbf{H}_T^{(S)},
\qquad
\mathbf{H}_I=\mathbf{H}_I^{(S)}.
\end{equation}
Since downstream classification requires fixed-length representations, each modality is summarized by attentive pooling. For modality \(M\in\{T,I\}\) with token matrix \(\mathbf{H}_M\in\mathbb{R}^{L_M\times d}\), the pooling weights are computed as
\begin{equation}
\label{eq:pooling_weights}
\boldsymbol{\alpha}_M
=
\mathrm{softmax}
\left(
\frac{\mathbf{H}_M\mathbf{u}_M}{\sqrt{d}}
\right),
\qquad
\mathbf{u}_M\in\mathbb{R}^{d},
\end{equation}
and the pooled representation is
\begin{equation}
\label{eq:pooled_representation}
\mathbf{z}_M
=
\sum_{m=1}^{L_M}
\alpha_{M,m}\mathbf{H}_{M,m}.
\end{equation}
The learnable query vector \(\mathbf{u}_M\) enables the model to assign different importance weights to different tokens or image patches instead of treating all positions uniformly.

\subsection{Adaptive Gated Fusion}

The pooled text and image vectors are fused through a dimension-wise gating mechanism. The gate is computed from the concatenated modality representations:
\begin{equation}
\label{eq:gate_computation}
\begin{aligned}
\mathbf{s}_g
&=
\mathbf{W}_{g,1}[\mathbf{z}_T;\mathbf{z}_I]
+
\mathbf{b}_{g,1},\\
\mathbf{g}
&=
\sigma
\left(
\mathbf{W}_{g,2}
\phi(\mathbf{s}_g)
+
\mathbf{b}_{g,2}
\right),
\qquad
\mathbf{g}\in(0,1)^d.
\end{aligned}
\end{equation}
where \(\phi(\cdot)\) is a nonlinear activation function, \(\sigma(\cdot)\) is the sigmoid function, and \([\cdot;\cdot]\) denotes vector concatenation. The fused classical representation is then obtained as
\begin{equation}
\label{eq:classical_fused_feature}
\mathbf{z}_C
=
\mathbf{g}\odot \mathbf{z}_T
+
(1-\mathbf{g})\odot \mathbf{z}_I,
\end{equation}
where \(\odot\) denotes element-wise multiplication. This formulation gives each latent dimension an adaptive modality preference. When \(g_j\) is close to one, the \(j\)-th fused feature relies more on the text representation; when \(g_j\) is close to zero, it relies more on the image representation.

\subsection{Parallel Quantum Feature Augmentation}

The fused classical vector $\mathbf{z}_C$ is passed to a set of $K$ parallel parameterized quantum circuits. 
Each circuit acts as a trainable quantum feature map that transforms the same fused representation into measurement readout features.

Let \(n\) denote the number of qubits and let \(D=2^n\) be the Hilbert space input dimension. In our implementation, \(D=d\), so the fused vector can be directly amplitude-encoded after normalization. We compute
\begin{equation}
\label{eq:normalized_classical_feature}
\widetilde{\mathbf{z}}_C
=
\frac{\mathbf{z}_C}
{\|\mathbf{z}_C\|_2+\epsilon},
\end{equation}
where \(\epsilon\) is a small constant for numerical stability. Amplitude encoding prepares the initial quantum state as
\begin{equation}
\label{eq:amplitude_encoding}
|\psi_0(\widetilde{\mathbf{z}}_C)\rangle
=
A(\widetilde{\mathbf{z}}_C)|0\rangle^{\otimes n}
=
\sum_{j=0}^{D-1}
\widetilde{z}_{C,j}|j\rangle .
\end{equation}
Here, \(A(\cdot)\) denotes the state-preparation operation and \(\{|j\rangle\}_{j=0}^{D-1}\) is the computational basis.

\begin{figure*}[t]
\centering
\includegraphics[width=\textwidth]{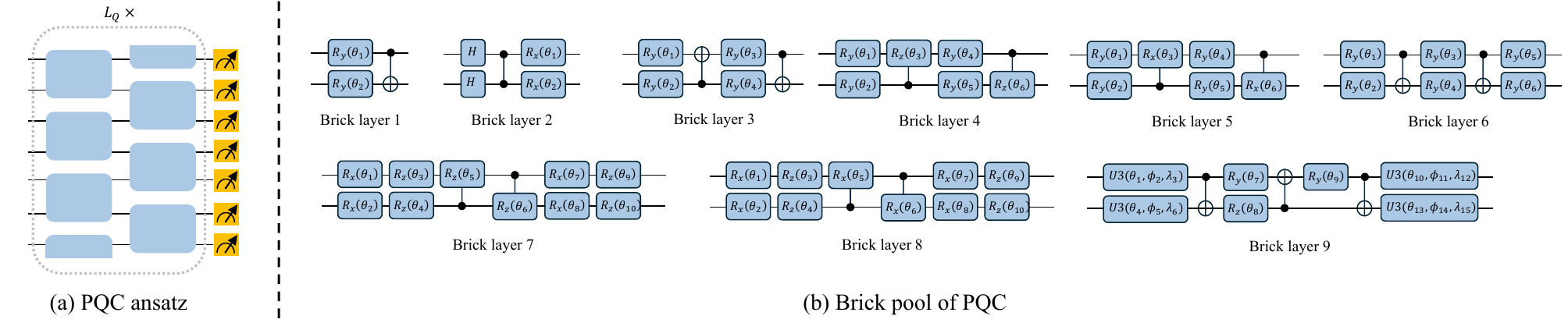}
\caption{
Schematic illustration of the Parallel Quantum Feature Augmentation module. 
(a) A shallow brickwall parameterized quantum circuit used in each quantum branch. 
(b) The predefined brick pool consisting of multiple local two-qubit parameterized blocks, which introduces structural diversity across parallel quantum branches.
}
\label{fig:pqc_ansatz_pool}
\end{figure*}

As illustrated in Fig.~\ref{fig:pqc_ansatz_pool}, each parallel quantum branch applies a shallow trainable brickwall circuit to the encoded state:
\begin{equation}
\label{eq:quantum_state_evolution}
|\psi_k(\widetilde{\mathbf{z}}_C;\boldsymbol{\theta}_k)\rangle
=
U_k(\boldsymbol{\theta}_k)
|\psi_0(\widetilde{\mathbf{z}}_C)\rangle,
\qquad
k=1,\ldots, K.
\end{equation}
The unitary transformation \(U_k(\boldsymbol{\theta}_k)\) is composed of local parameterized blocks arranged in a brickwall pattern:
\begin{equation}
\label{eq:unitary_decomposition}
U_k(\boldsymbol{\theta}_k)
=
\prod_{\ell=1}^{L_Q}
\prod_{(a,b)\in\mathcal{E}_{\ell}}
B_{b_k}^{(a,b)}
\left(
\boldsymbol{\theta}_{k,\ell,a,b}
\right),
\end{equation}
where \(L_Q\) is the number of quantum layers, \(\mathcal{E}_{\ell}\) denotes the set of qubit pairs activated at layer \(\ell\), and \(B_{b_k}^{(a,b)}(\cdot)\) is a two-qubit parameterized block selected from a predefined brick pool.

The brick pool is denoted by
\begin{equation}
\label{eq:brick_pool}
\mathcal{B}
=
\{B^{(1)},B^{(2)},\ldots,B^{(T)}\},
\end{equation}
where each \(B^{(t)}\) represents a parameterized two-qubit block with a specific arrangement of local rotations and entangling operations. 
For the \(k\)-th quantum branch, a brick type \(b_k\in\{1,\ldots,T\}\) is assigned from the brick pool and kept fixed during training, while its continuous parameters \(\boldsymbol{\theta}_k\) are optimized by backpropagation. 
Different branches may therefore implement different local circuit structures even though they receive the same fused input representation. 
This design introduces structural diversity across the parallel quantum feature maps without increasing the depth of any single branch. 
The purpose is to let the parallel readout branches produce complementary nonlinear transformations of the fused multimodal representation.

For each branch, we measure a set of observables:
\begin{equation}
\label{eq:observable_set}
\mathcal{O}=\{O_1,\ldots,O_m\}.
\end{equation}
The quantum feature vector of the \(k\)-th branch is
\begin{equation}
\label{eq:quantum_feature_vector}
\mathbf{q}_k
=
\left[
\langle O_1\rangle_k,
\ldots,
\langle O_m\rangle_k
\right],
\end{equation}
where each expectation value is computed as
\begin{equation}
\label{eq:expectation_value}
\langle O_r\rangle_k
=
\bigl\langle
\psi_k(\widetilde{\mathbf{z}}_C;\boldsymbol{\theta}_k)
\bigr|
O_r
\bigr|
\psi_k(\widetilde{\mathbf{z}}_C;\boldsymbol{\theta}_k)
\bigr\rangle .
\end{equation}

In this framework, a simple and hardware-compatible choice is to measure single-qubit Pauli-\(Z\) observables,
\begin{equation}
\label{eq:pauli_z_observables}
O_r
=
I^{\otimes(r-1)}
\otimes Z
\otimes I^{\otimes(n-r)},
\qquad r=1,\ldots,n .
\end{equation}
These measurements provide real-valued expectation readouts in \([-1,1]\). 
Since one Pauli-\(Z\) expectation value is measured from each qubit, each branch first yields a fixed-size readout vector \(\mathbf{q}_k\in[-1,1]^n\), where \(m=n\). 
In the implementation, this readout vector is passed through a lightweight branch-wise refinement layer before concatenation. 
For notational simplicity, \(\mathbf{q}_k\) denotes the resulting branch output in the following equations.
The outputs of all quantum branches are concatenated:
\begin{equation}
\label{eq:concatenated_quantum_feature}
\mathbf{q}
=
[\mathbf{q}_1;\mathbf{q}_2;\ldots;\mathbf{q}_K]
\in\mathbb{R}^{Km}.
\end{equation}

\subsection{Hybrid Representation and Prediction}

The final hybrid representation concatenates the classical fused feature and the quantum-augmented feature:
\begin{equation}
\label{eq:hybrid_representation}
\mathbf{h}
=
[\mathbf{z}_C;\mathbf{q}]
\in\mathbb{R}^{d+Km}.
\end{equation}
The classifier maps \(\mathbf{h}\) to logits:
\begin{equation}
\label{eq:logits}
\mathbf{o}
=
\mathrm{MLP}_{\omega}(\mathbf{h})
\in\mathbb{R}^{C}.
\end{equation}

For multi-label classification, the logits are converted into independent class probabilities:
\begin{equation}
\label{eq:multi_label_probabilities}
\mathbf{p}
=
\sigma(\mathbf{o}),
\end{equation}
and the predicted label vector is obtained using validation-selected thresholds:
\begin{equation}
\label{eq:multi_label_prediction}
\widehat{y}_c
=
\mathbb{I}(p_c\geq \tau_c),
\qquad
c=1,\ldots,C.
\end{equation}
The thresholds \(\{\tau_c\}_{c=1}^{C}\) are selected on the validation set and fixed before test evaluation to avoid test-set leakage.

For single-label classification, the logits are converted into a categorical distribution:
\begin{equation}
\label{eq:single_label_probabilities}
p_c
=
\frac{\exp(o_c)}
{\sum_{j=1}^{C}\exp(o_j)},
\end{equation}
and the predicted class is obtained by
\begin{equation}
\label{eq:single_label_prediction}
\widehat{y}
=
\arg\max_{c} p_c.
\end{equation}

\subsection{Training Objective}
\label{sec:training_objective}

The model is trained end-to-end with frozen unimodal encoders. The optimized parameters include the projection layers, cross-attention layers, pooling queries, gating module, quantum circuit parameters, and prediction head.

For multi-label classification, we use a class-wise sigmoid objective. The binary cross-entropy loss is written as
\begin{equation}
\label{eq:bce_loss}
\begin{aligned}
\mathcal{L}_{\mathrm{BCE}}
=
-\frac{1}{C}
\sum_{c=1}^{C}
\Big[
y_c\log(p_c+\epsilon) 
+
(1-y_c)\log(1-p_c+\epsilon)
\Big].
\end{aligned}
\end{equation}

When class imbalance is considered, this objective can be replaced by an asymmetric focal variant:
\begin{equation}
\label{eq:asl_loss}
\begin{aligned}
\mathcal{L}_{\mathrm{ASL}}
=
-\frac{1}{C}
\sum_{c=1}^{C}
\Big[
y_c(1-p_c)^{\gamma_+}\log(p_c+\epsilon)
+
(1-y_c)p_c^{\gamma_-}
\log(1-p_c+\epsilon)
\Big].
\end{aligned}
\end{equation}
where \(\gamma_+\) and \(\gamma_-\) control the focusing strength for positive and negative labels, respectively.

For single-label classification, we use the categorical cross-entropy loss:
\begin{equation}
\label{eq:ce_loss}
\mathcal{L}_{\mathrm{CE}}
=
-\log
\frac{\exp(o_{y})}
{\sum_{c=1}^{C}\exp(o_c)}.
\end{equation}

\subsection{Backpropagation and Hybrid Optimization}
\label{sec:backprop_optimization}

Algorithm~\ref{alg:pqfa} summarizes the training and validation procedure of PQFA. 
We now describe how the trainable classical modules and the quantum readout branch are optimized under the same supervised objective. 
The pretrained RoBERTa and ViT encoders remain frozen. 
The optimized parameter set is
\begin{equation}
\label{eq:optimized_parameters}
\boldsymbol{\Theta}
=
\{
\boldsymbol{\Theta}_{p},
\boldsymbol{\Theta}_{a},
\boldsymbol{\Theta}_{pool},
\boldsymbol{\Theta}_{g},
\boldsymbol{\Theta}_{q},
\boldsymbol{\omega}
\},
\end{equation}
where $\boldsymbol{\Theta}_{p}$, $\boldsymbol{\Theta}_{a}$, $\boldsymbol{\Theta}_{pool}$, and $\boldsymbol{\Theta}_{g}$ denote the parameters of the projection, cross-attention, attentive pooling, and gated fusion modules, respectively. 
The quantum circuit parameters are denoted by 
$\boldsymbol{\Theta}_{q}=\{\boldsymbol{\theta}_{1},\ldots,\boldsymbol{\theta}_{K}\}$, and $\boldsymbol{\omega}$ denotes the classifier parameters.

For a mini-batch 
$\mathcal{B}=\{(T_i,I_i,y_i)\}_{i=1}^{B}$, 
the empirical objective is
\begin{equation}
\label{eq:empirical_objective}
\mathcal{J}(\boldsymbol{\Theta})
=
\frac{1}{B}
\sum_{i=1}^{B}
\mathcal{L}(\mathbf{o}_i,y_i),
\end{equation}
where $\mathcal{L}$ is the task loss. 
For each sample, the logits are computed from the hybrid representation:
\begin{equation}
\label{eq:hybrid_logits_training}
\begin{aligned}
\mathbf{h}_i
&=
[
\mathbf{z}_{C,i};
\mathbf{q}_{i}
],\\
\mathbf{o}_i
&=
\mathrm{MLP}_{\boldsymbol{\omega}}(\mathbf{h}_i).
\end{aligned}
\end{equation}

The classifier gradient follows the usual chain rule:
\begin{equation}
\label{eq:classifier_gradient}
\frac{\partial \mathcal{J}}
{\partial \boldsymbol{\omega}}
=
\frac{1}{B}
\sum_{i=1}^{B}
\frac{\partial \mathcal{L}_i}
{\partial \mathbf{o}_i}
\frac{\partial \mathbf{o}_i}
{\partial \boldsymbol{\omega}}.
\end{equation}
More importantly, the fused representation receives gradients from two paths. 
Let
\begin{equation}
\label{eq:hybrid_gradient_split}
\boldsymbol{\delta}_{h,i}
=
\nabla_{\mathbf{h}_i}\mathcal{L}_i
=
[
\boldsymbol{\delta}_{C,i};
\boldsymbol{\delta}_{q,i}
],
\end{equation}
where $\boldsymbol{\delta}_{C,i}\in\mathbb{R}^{d}$ is the gradient associated with the direct classical part of $\mathbf{h}_i$, and 
$\boldsymbol{\delta}_{q,i}\in\mathbb{R}^{Km}$ is the gradient associated with the quantum readout part. 
Define
\begin{equation}
\label{eq:jacobian_definitions}
\begin{aligned}
\mathbf{J}_{\mathrm{norm},i}
&=
\frac{\partial \widetilde{\mathbf{z}}_{C,i}}
{\partial \mathbf{z}_{C,i}},\\
\mathbf{J}_{q,i}
&=
\frac{\partial \mathbf{q}_{i}}
{\partial \widetilde{\mathbf{z}}_{C,i}} .
\end{aligned}
\end{equation}
Then the total gradient received by the fused representation is
\begin{equation}
\label{eq:zc_gradient_compact}
\nabla_{\mathbf{z}_{C,i}}\mathcal{L}_i
=
\boldsymbol{\delta}_{C,i}
+
\mathbf{J}_{\mathrm{norm},i}^{\top}
\mathbf{J}_{q,i}^{\top}
\boldsymbol{\delta}_{q,i}.
\end{equation}
The first term comes from the direct classical path, while the second term comes from the quantum augmentation path.

The normalization before amplitude encoding is differentiable. 
For
\begin{equation}
\label{eq:amplitude_input_normalization}
\widetilde{\mathbf{z}}_{C}
=
\frac{\mathbf{z}_{C}}
{\|\mathbf{z}_{C}\|_{2}+\epsilon},
\end{equation}
with \(\rho=\|\mathbf{z}_{C}\|_2\), the Jacobian is
\begin{equation}
\label{eq:normalization_jacobian}
\mathbf{J}_{\mathrm{norm}}
=
\frac{1}{\rho+\epsilon}\mathbf{I}
-
\frac{
\mathbf{z}_{C}\mathbf{z}_{C}^{\top}
}{
\rho(\rho+\epsilon)^2
}.
\end{equation}
This Jacobian allows gradients from the quantum readout path to propagate from 
\(\widetilde{\mathbf{z}}_C\) back to the fused representation \(\mathbf{z}_C\). 
Therefore, the quantum branch contributes not only to the optimization of its own circuit parameters 
\(\boldsymbol{\theta}_k\), but also to the end-to-end training of the preceding fusion modules.

For the $k$-th quantum branch, the $r$-th readout is
\begin{equation}
\label{eq:quantum_measurement_training}
q_{k,r}
=
\bigl\langle
\psi_k(
\widetilde{\mathbf{z}}_{C};
\boldsymbol{\theta}_{k})
\bigr|
O_r
\bigr|
\psi_k(
\widetilde{\mathbf{z}}_{C};
\boldsymbol{\theta}_{k})
\bigr\rangle .
\end{equation}
The gradient with respect to a quantum parameter $\theta_{k,s}$ is
\begin{equation}
\label{eq:quantum_parameter_gradient}
\frac{\partial \mathcal{J}}
{\partial \theta_{k,s}}
=
\frac{1}{B}
\sum_{i=1}^{B}
\sum_{r=1}^{m}
\frac{\partial \mathcal{L}_i}
{\partial q_{i,k,r}}
\frac{\partial q_{i,k,r}}
{\partial \theta_{k,s}}.
\end{equation}
Since $O_r$ is Hermitian and independent of $\theta_{k,s}$, the derivative of an expectation value can be written as
\begin{equation}
\label{eq:expectation_derivative}
\frac{\partial q_{k,r}}
{\partial \theta_{k,s}}
=
2\,\operatorname{Re}
\bigl[
\bigl\langle
\partial_{\theta_{k,s}}\psi_k
\bigr|
O_r
\bigr|
\psi_k
\bigr\rangle
\bigr].
\end{equation}

In our implementation, the quantum circuits are built with PennyLane and connected to the PyTorch computation graph through the Torch interface. 
The experiments use the differentiable \texttt{default.qubit} state vector simulator. 
Under this simulator-based automatic differentiation setting, gradients are obtained for both the quantum circuit parameters and the amplitude encoding inputs, rather than through a finite-shot parameter shift procedure. 
Therefore, the quantum branch can affect the preceding fused representation as well as its own circuit parameters.

The parameters are updated by Adam. 
For a parameter group $\boldsymbol{\Theta}_{u}\subseteq\boldsymbol{\Theta}$, the update is written as
\begin{equation}
\label{eq:adam_update}
\boldsymbol{\Theta}_{u}^{(t+1)}
=
\mathrm{Adam}
\left(
\boldsymbol{\Theta}_{u}^{(t)},
\nabla_{\boldsymbol{\Theta}_{u}}\mathcal{J}^{(t)},
\eta_u
\right),
\end{equation}
where $t$ is the optimization step and $\eta_u$ is the learning rate of the corresponding parameter group. 
We use separate learning rates for the classical and quantum parameter groups. 
This optimization scheme trains the fusion modules, quantum readout branch, and classifier under a unified objective while keeping the pretrained encoders fixed.

\begin{algorithm*}[t]
\caption{Training and validation procedure of PQFA}
\label{alg:pqfa}
\begin{algorithmic}[1]

\Require Training set $\mathcal{D}_{\mathrm{train}}$, validation set $\mathcal{D}_{\mathrm{val}}$, frozen encoders $E_T,E_I$, number of quantum branches $K$, learning rates $\eta_{\mathrm{cls}},\eta_q$, task type
\Ensure Best trainable parameters $\boldsymbol{\Theta}^{*}$ and validation-selected thresholds $\boldsymbol{\tau}^{*}$ for multi-label classification

\State Initialize classical trainable parameters $\boldsymbol{\Theta}_{\mathrm{cls}}$ in projection layers, cross-attention, pooling, gated fusion, and classifier
\State Initialize quantum circuit parameters $\boldsymbol{\Theta}_{q}=\{\boldsymbol{\theta}_1,\ldots,\boldsymbol{\theta}_K\}$
\State Freeze the pretrained text encoder $E_T$ and image encoder $E_I$
\State Initialize the best validation score $s^{*}$

\For{each training epoch}
    \For{each mini-batch $(T,I,y)\subset\mathcal{D}_{\mathrm{train}}$}
        \State Extract frozen features: $\mathbf{X}_T\leftarrow E_T(T)$ and $\mathbf{X}_I\leftarrow E_I(I)$
        \State Construct fused representation: $\mathbf{z}_C\leftarrow F_{\mathrm{fuse}}(\mathbf{X}_T,\mathbf{X}_I;\boldsymbol{\Theta}_{\mathrm{cls}})$
        \State Generate quantum readout: $\mathbf{q}\leftarrow F_q(\mathbf{z}_C;\boldsymbol{\Theta}_q)$
        \State Form hybrid representation: $\mathbf{h}\leftarrow[\mathbf{z}_C;\mathbf{q}]$
        \State Compute logits and loss: $\mathbf{o}\leftarrow \mathrm{MLP}_{\boldsymbol{\omega}}(\mathbf{h})$, $\mathcal{L}\leftarrow\mathcal{L}(\mathbf{o},y)$
        \State Update $\boldsymbol{\Theta}_{\mathrm{cls}}$ and $\boldsymbol{\Theta}_{q}$ by Adam using separate learning rates $\eta_{\mathrm{cls}}$ and $\eta_q$
    \EndFor

    \State Evaluate the current model on $\mathcal{D}_{\mathrm{val}}$

    \If{the task is multi-label classification}
        \State Select class-wise thresholds $\boldsymbol{\tau}$ on $\mathcal{D}_{\mathrm{val}}$
        \State Compute the validation score using $\boldsymbol{\tau}$
    \Else
        \State Compute the validation score using the softmax prediction rule
    \EndIf

    \If{the validation score improves over $s^{*}$}
        \State Save $\boldsymbol{\Theta}^{*}\leftarrow\{\boldsymbol{\Theta}_{\mathrm{cls}},\boldsymbol{\Theta}_{q}\}$
        \State Save $\boldsymbol{\tau}^{*}\leftarrow\boldsymbol{\tau}$ for multi-label classification
        \State Update $s^{*}$
    \EndIf
\EndFor

\end{algorithmic}
\end{algorithm*}

\section{Experiments}
\label{sec:experiments}

\subsection{Experimental Setup}

\subsubsection{Datasets}

We evaluate PQFA on two image--text classification benchmarks with different prediction formats. This design allows us to examine whether the proposed post-fusion quantum augmentation strategy is effective in both multi-label and single-label multimodal classification settings.

\paragraph{MM-IMDb}
MM-IMDb is a multimodal movie genre classification benchmark~\cite{arevalo2020gated}. Each sample contains a movie poster, a plot summary, and one or more genre labels. Since a movie can be associated with multiple genres, MM-IMDb is formulated as a multi-label classification task with 23 categories. After preprocessing and filtering samples with valid image--text pairs, the controlled test split used in our experiments contains 3,894 samples.

\paragraph{N24News}
N24News is a multimodal news classification benchmark constructed from New York Times articles and associated images~\cite{wang2022n24news}. Each sample contains a news image, a textual article field, and a single topic label from 24 categories. In this work, we use the abstract field as the textual input and the corresponding news image as the visual input. Since each sample is assigned to exactly one category, N24News is treated as a single-label classification task.

\subsubsection{Backbone Encoders and Training Protocol}

We use frozen pretrained Transformer encoders for both modalities. 
The text stream follows RoBERTa~\cite{liu2019roberta}, a robustly optimized BERT variant for contextual language representation, and the image stream follows the Vision Transformer~\cite{dosovitskiy2021an}, which represents an image as a sequence of patch tokens. 
The encoder parameters are fixed throughout training, and only the projection layers, cross-attention modules, attentive pooling, gated fusion module, augmentation branch, and classifier are optimized. 
This protocol keeps all compared models under the same encoder and fusion backbone, so that the comparison focuses on the effect of feature augmentation.

Images are resized to $224\times224$, and text sequences are truncated to 256 tokens. 
Both modalities are projected into a 128-dimensional latent space. 
For the quantum branch, the fused representation is normalized and amplitude-encoded into a quantum state with \(n=7\) qubits, satisfying \(2^n=128\). Each quantum branch produces \(m=n\) Pauli-\(Z\) expectation readouts. The number of parallel quantum branches \(K\) is selected according to validation performance and then fixed within each controlled comparison. Since each branch produces \(m\) readouts, the quantum augmentation vector has dimension \(Km\). The width-matched MLP-Aug baseline is constructed with the same augmentation output dimension \(Km\), ensuring that the comparison between classical and quantum augmentation is not confounded by feature dimensionality.

The hybrid models are implemented with PennyLane and PyTorch. 
Quantum circuits are simulated on the differentiable backend and connected to the PyTorch computation graph through the Torch interface. 
Classical and quantum parameter groups are optimized by Adam with learning rates $1\times10^{-4}$ and $1\times10^{-5}$, respectively, and the batch size is 16. 
Model selection is performed on the validation set. 
For multi-label classification, class-wise thresholds are selected on the validation split and fixed before test evaluation to avoid test set leakage.

Unless otherwise specified, all controlled experiments are repeated with five independent random seeds, and the reported results are averaged across runs.

\subsection{Evaluation Metrics and Statistical Diagnostics}
\label{sec:eval_metrics}

We use different evaluation metrics for the multi-label and single-label classification settings. 
For MM-IMDb, each sample may be associated with multiple genre labels, and we therefore report Micro-F1 and Macro-F1 as the primary metrics~\cite{powers2011evaluation}. 
For N24News, each sample belongs to exactly one category, and we report Accuracy as the main metric.

For multi-label classification, let $TP_c$, $FP_c$, and $FN_c$ denote the number of true positives, false positives, and false negatives for class $c$, respectively. 
The class-wise precision and recall are defined as
\begin{equation}
P_c = \frac{TP_c}{TP_c+FP_c},
\qquad
R_c = \frac{TP_c}{TP_c+FN_c}.
\end{equation}
The class-wise F1 score is then computed as
\begin{equation}
F1_c =
\frac{2P_cR_c}{P_c+R_c}
=
\frac{2TP_c}{2TP_c+FP_c+FN_c}.
\end{equation}
Macro-F1 gives equal weight to each class and is defined as
\begin{equation}
\mathrm{Macro\text{-}F1}
=
\frac{1}{C}
\sum_{c=1}^{C}
F1_c.
\end{equation}
This metric is sensitive to minority categories because it averages the F1 scores over classes rather than over individual label decisions.

Micro-F1 first aggregates the true positives, false positives, and false negatives over all classes, and then computes the F1 score:
\begin{equation}
\mathrm{Micro\text{-}F1}
=
\frac{
2\sum_{c=1}^{C}TP_c
}{
2\sum_{c=1}^{C}TP_c
+
\sum_{c=1}^{C}FP_c
+
\sum_{c=1}^{C}FN_c
}.
\end{equation}
Compared with Macro-F1, Micro-F1 reflects the overall decision quality across all instance-label pairs and is more influenced by frequent classes.

For single-label classification, we report Accuracy:
\begin{equation}
\mathrm{Accuracy}
=
\frac{1}{N}
\sum_{i=1}^{N}
\mathbb{I}(\hat{y}_i=y_i),
\end{equation}
where $\hat{y}_i$ and $y_i$ denote the predicted and ground-truth class labels of the $i$-th sample, respectively, and $\mathbb{I}(\cdot)$ is the indicator function.

\subsection{Reference Benchmarking}
\label{sec:reference_benchmarking}
Reference benchmarking compares PQFA with representative results reported in previous studies. 

\subsubsection{Results on MM-IMDb}

\begin{table*}[t]
\centering
\caption{Reference comparison on MM-IMDb.}
\label{tab:mmimdb_main}

\setlength{\tabcolsep}{4pt}
\begin{tabular}{lcccc}
\toprule
Model & Image Encoder & Text Encoder & Micro-F1 & Macro-F1 \\
\midrule
LRMF \cite{liu2018efficient} & ResNet & GloVe & 58.95 & 50.73 \\
MFM \cite{tsai2019learning} & ResNet & BERT & 56.44 & 48.53 \\
MI-Matrix \cite{jayakumar2020multiplicative} & ResNet & BERT & 55.87 & 46.77 \\
RMFE \cite{gat2020removing} & ResNet & BERT & 58.67 & 49.82 \\
CCA \cite{sun2020learning} & ResNet & BERT & 60.31 & 50.45 \\
RefNet \cite{sankaran2021multimodal} & ResNet & BERT & 59.45 & 51.51 \\
DynMM \cite{xue2023dynamic} & ResNet & BERT & 60.35 & 51.60 \\
ModiFedCat \cite{li2025modifedcat} & ResNet & BERT & 61.14 & -- \\
Coupled Mamba \cite{li2024coupled} & ResNet & BERT & 62.41 & 52.58 \\
InfMasking \cite{wen2026infmasking} & ResNet & BERT & -- & 55.93 \\
MBT \cite{nagrani2021attention} & ViT & BERT & 64.81 & 59.60 \\
\textbf{PQFA (Ours)} & ViT & RoBERTa & \textbf{68.28} & \textbf{61.85} \\
\bottomrule
\end{tabular}
\end{table*}

Table~\ref{tab:mmimdb_main} compares PQFA with representative multimodal fusion methods on MM-IMDb. The compared methods cover several major technical routes, including factorized or multiplicative fusion, robust multimodal representation learning, cross-modal alignment, dynamic fusion, sequence modeling, information masking, and Transformer-based multimodal fusion. Specifically, LRMF and MI-Matrix model high-order multimodal interactions through factorized or multiplicative representations \cite{liu2018efficient,jayakumar2020multiplicative}; MFM and RMFE focus on robust multimodal representation learning \cite{tsai2019learning,gat2020removing}; CCA and RefNet improve cross-modal alignment and refinement \cite{sun2020learning,sankaran2021multimodal}; DynMM and ModiFedCat introduce dynamic or adaptive fusion mechanisms \cite{xue2023dynamic,li2025modifedcat}; Coupled Mamba and InfMasking represent recent sequence-modeling and information-masking based approaches \cite{li2024coupled,wen2026infmasking}; and MBT provides a strong Transformer-based multimodal fusion reference \cite{nagrani2021attention}.

PQFA achieves 68.28 Micro-F1 and 61.85 Macro-F1, which are higher than the listed reference results under this reporting setup. 
The Macro-F1 result is particularly relevant because MM-IMDb is label-imbalanced, and Macro-F1 better reflects category-balanced recognition. 
However, because Table~\ref{tab:mmimdb_main} includes methods with different encoder choices and training settings, it should be interpreted as a reference comparison rather than as direct evidence that the quantum augmentation branch alone causes the improvement. 
The controlled ablation study in Section~\ref{sec:ablation} provides the primary evidence for isolating the contribution of the proposed quantum augmentation branch.

\subsubsection{Results on N24News}

\begin{table}[t]
\centering
\small
\caption{Reference comparison on N24News.}
\label{tab:n24news_main}
\setlength{\tabcolsep}{4pt}
\begin{tabular}{lccc}
\toprule
Model & Image Encoder & Text Encoder & Acc.(\%) \\
\midrule
N24News \cite{wang2022n24news} & ViT & RoBERTa & 83.33 \\
UniS-MMC \cite{zou2023unis} & ViT & BERT & 84.20 \\
SDDA \cite{chen2025sdda} & ViT & RoBERTa & 84.43 \\
M3CoL \cite{kumar2025m3col} & ViT & RoBERTa & 84.70 \\
\textbf{PQFA (Ours)} & ViT & RoBERTa & \textbf{85.35} \\
\bottomrule
\end{tabular}
\end{table}

Table~\ref{tab:n24news_main} reports the reference comparison on N24News. The compared methods include the official N24News multimodal baseline and recent image--text classification approaches based on contrastive learning, self-distillation, and multimodal mixup. The official N24News baseline evaluates multimodal news classification with pretrained visual and textual encoders, showing that news images provide complementary information beyond text-only classification \cite{wang2022n24news}. UniS-MMC introduces unimodality-supervised multimodal contrastive learning, where unimodal predictions are used to guide cross-modal representation alignment \cite{zou2023unis}. SDDA employs progressive self-distillation with decoupled alignment to alleviate modality heterogeneity and improve image--text representation learning \cite{chen2025sdda}. M3CoL explores multimodal mixup contrastive learning, aiming to capture shared relations across modalities through mixup-based contrastive supervision and auxiliary unimodal prediction modules \cite{kumar2025m3col}.

PQFA achieves 85.35\% accuracy on N24News, which is higher than the listed reference results. 
As with MM-IMDb, this table is mainly used to position PQFA relative to existing image--text classification methods, while the controlled ablation in Section~\ref{sec:ablation} is used for attribution. 
The result indicates that the proposed post-fusion quantum augmentation framework is not restricted to multi-label movie-genre classification, but can also be instantiated for single-label multimodal news classification by changing the prediction head and task loss.

\subsection{Controlled Ablation and Parameter Efficiency}
\label{sec:ablation}

The reference comparisons position PQFA against existing multimodal methods, but they do not by themselves identify which component is responsible for the gain. We therefore conduct controlled ablations in which the frozen encoders, data splits, projection dimension, and training protocol for the shared classical components are kept fixed. Under this protocol, the comparisons focus on the fusion strategy and on the post-fusion augmentation branch.

The ablation is organized around three questions. First, text-only and image-only variants test whether the two modalities provide complementary information. Second, average fusion, concatenation fusion, and gated fusion compare fixed fusion rules with adaptive feature-level routing. Third, NoQ and MLP-Aug isolate the contribution of the quantum branch. NoQ denotes the gated-fusion backbone without an augmentation branch. MLP-Aug replaces the quantum readout with a classical feed-forward branch of the same augmentation output dimension, so that the comparison with PQFA is controlled for the final augmentation width.

\begin{table}[t]
\centering
\small
\caption{Controlled ablation results on MM-IMDb and N24News. All variants use the same frozen encoders, data split, projection dimension, and training protocol for the shared classical components.}
\label{tab:controlled_ablation}
\setlength{\tabcolsep}{4pt}
\begin{tabular}{lccc}
\toprule
\multirow{2}{*}{Method} & \multicolumn{2}{c}{MM-IMDb} & N24News \\
\cmidrule(lr){2-3}\cmidrule(lr){4-4}
 & Micro-F1 & Macro-F1 & Acc.(\%) \\
\midrule
Text only        & 63.88 & 53.12 & 80.64 \\
Image only       & 32.63 & 14.36 & 55.28 \\
Average fusion   & 66.87 & 59.77 & 83.47 \\
Concat fusion    & 67.09 & 60.18 & 83.72 \\
NoQ              & 67.57 & 60.98 & 84.70 \\
MLP-Aug          & 67.62 & 61.22 & 84.23 \\
\textbf{PQFA}    & \textbf{68.28} & \textbf{61.85} & \textbf{85.35} \\
\bottomrule
\end{tabular}
\end{table}

Table~\ref{tab:controlled_ablation} gives the main ablation results. The two single-modality variants confirm that both modalities contribute useful information, although the text stream is substantially stronger than the image stream on MM-IMDb. Among the fusion baselines, the gated NoQ backbone improves over average and concatenation fusion, indicating that adaptive feature-wise routing is more effective than fixed combination rules. PQFA achieves the best performance on both benchmarks.

The comparison with MLP-Aug is the most direct controlled test of the quantum augmentation branch. On MM-IMDb, MLP-Aug only slightly improves over NoQ, increasing Micro-F1 from 67.57 to 67.62 and Macro-F1 from 60.98 to 61.22. PQFA further improves the scores to 68.28 Micro-F1 and 61.85 Macro-F1. On N24News, MLP-Aug does not improve over NoQ, whereas PQFA increases accuracy from 84.70\% to 85.35\%. These results indicate that a width-matched classical augmentation branch is not sufficient to reproduce the gain obtained by the proposed quantum readout branch.

\begin{table}[t]
\centering
\small
\caption{Trainable parameter counts of the full PQFA model and the augmentation branches. Frozen RoBERTa and ViT encoder parameters are excluded.}
\label{tab:parameter_count}
\setlength{\tabcolsep}{6pt}
\begin{tabular}{lc}
\toprule
Component & Trainable parameters \\
\midrule
Full PQFA model & 782.9K \\
PQFA augmentation branch & 2.2K \\
MLP-Aug branch & 24.0K \\
\bottomrule
\end{tabular}
\end{table}

Table~\ref{tab:parameter_count} contextualizes the same comparison from the perspective of trainable parameter count. The full PQFA model contains 782.9K trainable parameters, excluding the frozen RoBERTa and ViT encoders. The PQFA augmentation branch accounts for only 2.2K parameters, including 1,596 quantum circuit parameters and the lightweight readout refinement layers following quantum measurement. This is approximately 0.28\% of the trainable parameters of the full PQFA model. By contrast, the width-matched MLP-Aug branch contains 24.0K trainable parameters. Hence, the improvement of PQFA is not attributable to using a larger augmentation branch. Rather, the results suggest that a small trainable quantum-enhanced branch can provide an additional post-fusion transformation on top of a much larger classical fusion backbone.

We therefore frame the ablation result as evidence for parameter-efficient post-fusion augmentation rather than as a hardware-level quantum advantage claim. Under the same encoder, fusion, and feature-width constraints, the proposed parallel quantum readout provides an effective augmentation module with substantially fewer branch parameters than the width-matched classical alternative.

\subsection{Robustness under Missing Modalities}
\label{sec:missing_modality}

The controlled ablation above evaluates complete-input performance. In practical multimodal systems, however, textual fields or images may be unavailable at inference time. We therefore evaluate whether the advantage of PQFA persists under controlled missing-modality perturbations on MM-IMDb. Given a missing rate $p$, a proportion $p$ of the test samples is randomly selected, and one modality is replaced by a null input while the total test set size remains unchanged. For the image-missing condition, the image tensor is replaced by a zero tensor after preprocessing. For the text-missing condition, the textual input is replaced by an empty string and processed by the same tokenizer. NoQ, MLP-Aug, and PQFA are evaluated under identical missing masks.

\begin{table*}[t]
\centering
\caption{Performance comparison under different missing-modality settings and missing rates. Reported values are averaged over five random seeds.}
\label{tab:missing_modality_results}
\begin{tabular}{ll cc cc cc}
\toprule
\multirow{2}{*}{\textbf{Missing Modality}} & \multirow{2}{*}{\textbf{Rate}} & \multicolumn{2}{c}{\textbf{NoQ}} & \multicolumn{2}{c}{\textbf{MLP-Aug}} & \multicolumn{2}{c}{\textbf{PQFA}} \\
\cmidrule(lr){3-4} \cmidrule(lr){5-6} \cmidrule(lr){7-8}
& & \textbf{Micro-F1} & \textbf{Macro-F1} & \textbf{Micro-F1} & \textbf{Macro-F1} & \textbf{Micro-F1} & \textbf{Macro-F1} \\
\midrule
Clean & 0.0 & 67.57 & 60.98 & 67.62 & 61.22 & \textbf{68.28} & \textbf{61.85} \\
\midrule
\multirow{5}{*}{Image Missing}
& 0.1 & 67.15 & 60.23 & 67.20 & 60.44 & \textbf{67.75} & \textbf{60.80} \\
& 0.3 & 66.20 & 59.56 & 66.22 & 59.20 & \textbf{66.76} & \textbf{59.60} \\
& 0.5 & 65.25 & 58.09 & 65.92 & 57.66 & \textbf{66.09} & \textbf{58.23} \\
& 0.7 & 64.28 & 55.93 & 64.80 & 55.94 & \textbf{65.12} & \textbf{56.70} \\
& 0.9 & 63.35 & 55.02 & 64.17 & 53.76 & \textbf{64.19} & \textbf{55.07} \\
\midrule
\multirow{5}{*}{Text Missing}
& 0.1 & 64.20 & 57.23 & 64.52 & 57.45 & \textbf{65.57} & \textbf{58.44} \\
& 0.3 & 57.66 & 51.16 & 58.02 & 50.61 & \textbf{59.25} & \textbf{51.28} \\
& 0.5 & 50.72 & 40.74 & 51.02 & 41.57 & \textbf{52.42} & \textbf{41.60} \\
& 0.7 & 39.52 & 31.76 & 43.25 & 32.48 & \textbf{45.31} & \textbf{32.49} \\
& 0.9 & 27.78 & 16.12 & 34.51 & 18.54 & \textbf{36.96} & \textbf{19.62} \\
\bottomrule
\end{tabular}
\end{table*}

Table~\ref{tab:missing_modality_results} reports the robustness results. PQFA achieves the best Micro-F1 and Macro-F1 in every reported missing-modality setting. The advantage is especially pronounced when the text modality is removed. When the text-missing rate increases to $p=0.9$, NoQ drops to 27.78 Micro-F1, MLP-Aug reaches 34.51, and PQFA further improves to 36.96. This result indicates that the quantum augmentation branch provides useful post-fusion features when the dominant textual signal is unavailable and the model must rely more heavily on the remaining visual information.

The image-missing setting shows a milder performance decline, indicating that MM-IMDb relies more heavily on text than on poster images. This is also consistent with the single-modality results in Table~\ref{tab:controlled_ablation}, where the text-only model substantially outperforms the image-only model. Even in this relatively less challenging setting, PQFA maintains the leading performance across all image-missing rates. For example, when \(p=0.9\), PQFA obtains 64.19 Micro-F1, which is slightly higher than MLP-Aug and clearly higher than NoQ.

The text-missing setting provides a more stringent robustness test because it removes the stronger modality. Under this condition, the benefit of PQFA becomes larger. At \(p=0.7\), PQFA improves Micro-F1 from 39.52 with NoQ and 43.25 with MLP-Aug to 45.31. At \(p=0.9\), PQFA improves over MLP-Aug by 2.45 Micro-F1 and over NoQ by 9.18 Micro-F1. These results show that the proposed quantum readout branch is particularly effective when the fused representation is strongly perturbed by the loss of the dominant modality.

Overall, the missing-modality experiment provides one of the strongest pieces of evidence for PQFA. The model not only improves complete-input performance, but also consistently preserves stronger predictive performance under controlled image-missing and text-missing perturbations. This supports the use of parallel quantum readouts as a robust post-fusion augmentation mechanism for incomplete multimodal inputs.

\subsection{Paired Error Transition and Label-level Analysis}
\label{sec:error_transition}

The missing-modality experiment evaluates robustness at the aggregate level. We next inspect clean-test predictions at a finer resolution. Aggregate metrics indicate whether a model improves overall performance, but they do not reveal how individual predictions change. 
To examine whether PQFA produces systematic corrections rather than merely shifting aggregate scores, we conduct a paired error transition analysis on the MM-IMDb test set. 
Since MM-IMDb is a multi-label task, each sample-label pair is treated as an individual binary decision.

We first use the sample-wise Jaccard score to measure the overlap between the predicted and ground-truth label sets. 
Let $\hat{Y}_i$ and $Y_i$ denote the predicted and ground-truth label sets of the $i$-th sample. 
The Jaccard score is defined as
\begin{equation}
J_i
=
\frac{
|\hat{Y}_i \cap Y_i|
}{
|\hat{Y}_i \cup Y_i|
}.
\end{equation}
When comparing a baseline model $A$ with PQFA, denoted as model $B$, we report the mean sample-wise Jaccard change:
\begin{equation}
\Delta \bar{J}
=
\frac{1}{N}
\sum_{i=1}^{N}
\left(
J_i^{B} - J_i^{A}
\right).
\end{equation}
A positive $\Delta \bar{J}$ indicates that PQFA improves the average overlap between predicted and ground-truth label sets.

We further define two paired decision transition sets. 
A corrected decision is a sample-label pair that is incorrectly predicted by the baseline but correctly predicted by PQFA:
\begin{equation}
\mathcal{S}_{\mathrm{corr}}
=
\{(i,c) \mid \hat{y}^{A}_{ic}\neq y_{ic},\;
\hat{y}^{B}_{ic}=y_{ic}\}.
\end{equation}
A corrupted decision is a sample-label pair that is correctly predicted by the baseline but incorrectly predicted by PQFA:
\begin{equation}
\mathcal{S}_{\mathrm{corrup}}
=
\{(i,c) \mid \hat{y}^{A}_{ic}=y_{ic},\;
\hat{y}^{B}_{ic}\neq y_{ic}\}.
\end{equation}
The net corrected decisions are computed as
\begin{equation}
\Delta N
=
|\mathcal{S}_{\mathrm{corr}}|
-
|\mathcal{S}_{\mathrm{corrup}}|.
\end{equation}
We also define the Net Correction Ratio as an auxiliary paired diagnostic statistic:
\begin{equation}
\mathrm{NCR}
=
\frac{
|\mathcal{S}_{\mathrm{corr}}|
-
|\mathcal{S}_{\mathrm{corrup}}|
}{
|\mathcal{S}_{\mathrm{corr}}|
+
|\mathcal{S}_{\mathrm{corrup}}|
}.
\end{equation}
A positive NCR indicates that PQFA corrects more decisions than it corrupts relative to the compared baseline.

For paired significance diagnostics, we use McNemar's test on discordant decisions~\cite{mcnemar1947note}. 
Let
\begin{equation}
n_{01}=|\mathcal{S}_{\mathrm{corr}}|,
\qquad
n_{10}=|\mathcal{S}_{\mathrm{corrup}}|.
\end{equation}
The continuity-corrected McNemar statistic is
\begin{equation}
\chi^2_{\mathrm{McN}}
=
\frac{
(|n_{01}-n_{10}|-1)^2
}{
n_{01}+n_{10}
},
\end{equation}
which approximately follows a chi-square distribution with one degree of freedom under the null hypothesis that the two models have the same error rate on paired decisions. 
In addition, we estimate the uncertainty of $\Delta \bar{J}$ using sample-level bootstrap confidence intervals~\cite{tibshirani1993introduction}. 
Specifically, we resample the test samples with replacement, recompute $\Delta \bar{J}$ for each bootstrap replicate, and report the percentile-based $95\%$ confidence interval.

\begin{table*}[t]
\centering
\caption{Paired decision-level transition between baseline models and PQFA on MM-IMDb.}
\label{tab:paired_transition_compact}
\setlength{\tabcolsep}{5pt}
\begin{tabular}{lcc}
\toprule
Statistic & PQFA vs NoQ & PQFA vs MLP-Aug \\
\midrule
Corrected decisions & 1,517 & 1,518 \\
Corrupted decisions & 1,444 & 1,337 \\
Net corrected decisions & +73 & +181 \\
Net correction ratio & 0.0247 & 0.0634 \\
Mean $\Delta J_i$ & +0.0075 & +0.0116 \\
Bootstrap 95\% CI & [0.0015, 0.0137] & [0.0057, 0.0177] \\
McNemar diagnostic $p$ & 0.1858 & $<0.001$ \\
\bottomrule
\end{tabular}
\end{table*}

Table~\ref{tab:paired_transition_compact} summarizes the decision-level transitions. 
Compared with MLP-Aug, PQFA corrects 1,518 erroneous decisions while introducing 1,337 new errors, resulting in a positive net correction of 181 decisions. 
The McNemar diagnostic gives $p<0.001$, and the bootstrap 95\% confidence interval of the mean sample-wise Jaccard gain, $[0.0057,0.0177]$, excludes zero. 
This indicates that the improvement over the classical augmentation baseline with the same output width is not only visible in aggregate F1 scores, but is also reflected in paired decision changes.

Compared with NoQ, PQFA also produces a positive but more moderate paired transition. 
It corrects 1,517 decisions and corrupts 1,444 decisions, yielding a net correction of 73 decisions and a positive NCR of 0.0247. 
The McNemar diagnostic is not significant at the conventional 0.05 level, suggesting that PQFA does not radically alter the individual binary decisions of the already strong gated-fusion backbone. 
However, the mean sample-wise Jaccard gain is positive, $\Delta \bar{J}=0.0075$, and its bootstrap 95\% confidence interval, $[0.0015,0.0137]$, excludes zero. 
This indicates that although the decision-slot correction imbalance over NoQ is modest, the sample-level label-set overlap still improves consistently.

Overall, these paired diagnostics provide a more fine-grained view of the controlled ablation results. 
The comparison with NoQ shows that the quantum branch improves the fused backbone without causing large disruptive prediction changes. 
The comparison with MLP-Aug shows that the same output-width classical augmentation does not reproduce the paired correction pattern of PQFA. 
Together, these results support the interpretation that the proposed quantum readout branch acts as a controlled post-fusion augmentation module rather than a generic feature-width expansion mechanism.

\begin{figure*}[t]
\centering
\includegraphics[width=\textwidth]{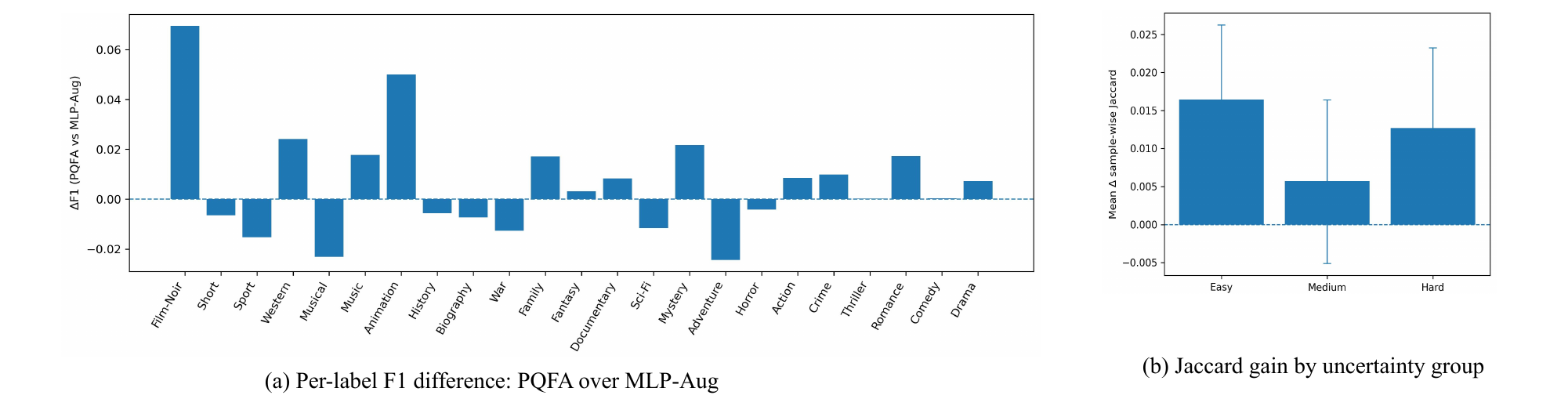}
\caption{
Fine-grained comparison between PQFA and MLP-Aug on MM-IMDb.
(a) Per-label F1 difference across genres sorted by support.
(b) Mean sample-wise Jaccard gain grouped by MLP-Aug prediction uncertainty.
Positive values indicate that PQFA outperforms the width-matched classical augmentation baseline.
}
\label{fig:transition_figs}
\end{figure*}

\begin{table*}[t]
\centering
\caption{Representative improved and degraded genres under PQFA compared with MLP-Aug.}
\label{tab:representative_labels}
\setlength{\tabcolsep}{4pt}
\begin{tabular}{llcccc}
\toprule
Type & Genre & Support & MLP-Aug F1 & PQFA F1 & $\Delta$F1 \\
\midrule
Improved & Film-Noir & 52 & 28.95 & \textbf{35.90} & +6.95 \\
Improved & Animation & 149 & 66.67 & \textbf{71.67} & +5.01 \\
Improved & Western & 114 & 74.14 & \textbf{76.56} & +2.42 \\
Degraded & Adventure & 395 & \textbf{59.42} & 56.98 & -2.43 \\
Degraded & Musical & 131 & \textbf{38.77} & 36.45 & -2.32 \\
Degraded & Sport & 103 & \textbf{72.56} & 71.03 & -1.53 \\
\bottomrule
\end{tabular}
\end{table*}

Fig.~\ref{fig:transition_figs} and Table~\ref{tab:representative_labels} further localize the transition behavior. 
PQFA improves the $F_1$ scores of 14 out of 23 genres compared with MLP-Aug. 
The largest gains occur in Film-Noir, Animation, and Western, while several categories show smaller or negative changes. 
This label-level variation indicates that PQFA does not simply apply a uniform score shift across all classes. 
Instead, the quantum readout branch produces category-sensitive post-fusion transformations, with stronger benefits in semantic regions where the MLP augmentation baseline leaves more correctable errors.

The sample-level uncertainty analysis in Fig.~\ref{fig:transition_figs}(b) shows positive mean Jaccard gains across easy, medium, and hard sample groups. 
This suggests that the corrective effect of PQFA is not confined to a single difficulty group, but appears across different levels of prediction uncertainty. 
Together with the label-level results, this pattern is more consistent with a structured post-fusion transformation than with a random or purely dimensional expansion of the feature space.

\subsection{Sensitivity to the Number of Quantum Branches}
\label{sec:k_sensitivity}

The preceding sections establish the main predictive and robustness effects of PQFA under the default branch configuration. We next examine how sensitive the method is to the number of parallel quantum branches on MM-IMDb. The branch number \(K\) controls the dimensionality of the quantum augmentation vector because each seven-qubit branch produces seven Pauli-\(Z\) expectation readouts. Increasing \(K\) can therefore provide a richer set of quantum feature transformations, but it can also introduce redundant or noisy readout dimensions.

\begin{table}[t]
\centering
\caption{Sensitivity analysis of the number of parallel quantum branches on MM-IMDb.}
\label{tab:k_ablation_mmimdb}
\begin{tabular}{ccc}
\toprule
Number of branches & Micro-F1 & Macro-F1  \\
\midrule
1 & 68.04 & \textbf{62.08} \\
2 & 67.98 & 62.02 \\
3 & 68.08 & 61.31 \\
4 & 68.03 & 61.74 \\
5 & 67.29 & 60.83 \\
6 & 68.21 & 61.90 \\
7 & 67.56 & 61.01 \\
8 & \textbf{68.28} & 61.85 \\
9 & 67.47 & 60.85 \\
\bottomrule
\end{tabular}
\end{table}

Table~\ref{tab:k_ablation_mmimdb} shows that performance is not monotonic in \(K\). A small number of branches already yields competitive results, indicating that useful post-fusion transformations can be obtained without a large parallel readout bank. The best Micro-F1 is obtained at \(K=8\), while the corresponding Macro-F1 remains close to the best observed value. Increasing the number of branches to \(K=9\) decreases both metrics, suggesting that additional branches do not automatically improve generalization. We therefore use \(K=8\) as the main MM-IMDb setting and regard it as an empirical trade-off between augmentation capacity and generalization rather than as evidence of monotonic scaling.

\subsection{Feature-space Diagnostics of Augmentation Variants}
\label{sec:feature_space_diagnostics}

The preceding experiments evaluate prediction accuracy, robustness, and branch-number sensitivity. We now examine whether the gain of PQFA is merely a consequence of feature-width expansion or whether the augmentation branch forms a more structured feature space. To this end, we analyze the intrinsic geometry of the augmentation features on MM-IMDb. This diagnostic focuses on the output features generated by different augmentation branches and is complementary to the aggregate performance comparison in Table~\ref{tab:controlled_ablation}.

For each trained model and each random seed, let $Q^{(s)} \in \mathbb{R}^{N \times d_a}$ denote the augmentation feature matrix extracted from the test set, where $N$ is the number of test samples and $d_a$ is the augmentation-feature dimension. We first center $Q^{(s)}$ and compute its covariance matrix:
\begin{equation}
\Sigma_Q^{(s)} = \frac{1}{N-1}\bar{Q}^{(s)\top}\bar{Q}^{(s)}.
\end{equation}
Let $\lambda_1^{(s)} \geq \lambda_2^{(s)} \geq \cdots \geq \lambda_{d_a}^{(s)}$ denote the eigenvalues of $\Sigma_Q^{(s)}$. The cumulative explained variance ratio of the first $k$ principal components is defined as
\begin{equation}
R^{(s)}(k)=
\frac{\sum_{j=1}^{k}\lambda_j^{(s)}}
{\sum_{j=1}^{d_a}\lambda_j^{(s)}}.
\end{equation}
We also compute the PCA-based effective dimension:
\begin{equation}
d_{\mathrm{eff}}^{(s)} =
\frac{
\left(\sum_{j=1}^{d_a}\lambda_j^{(s)}\right)^2
}{
\sum_{j=1}^{d_a}\left(\lambda_j^{(s)}\right)^2
}.
\end{equation}
In addition, $k_{90}$ and $k_{95}$ denote the minimum numbers of principal components required to explain 90\% and 95\% of the augmentation-feature variance, respectively. These quantities are computed separately for each seed and then averaged across random seeds.

\begin{figure}[t]
\centering
\includegraphics[width=0.7\columnwidth]{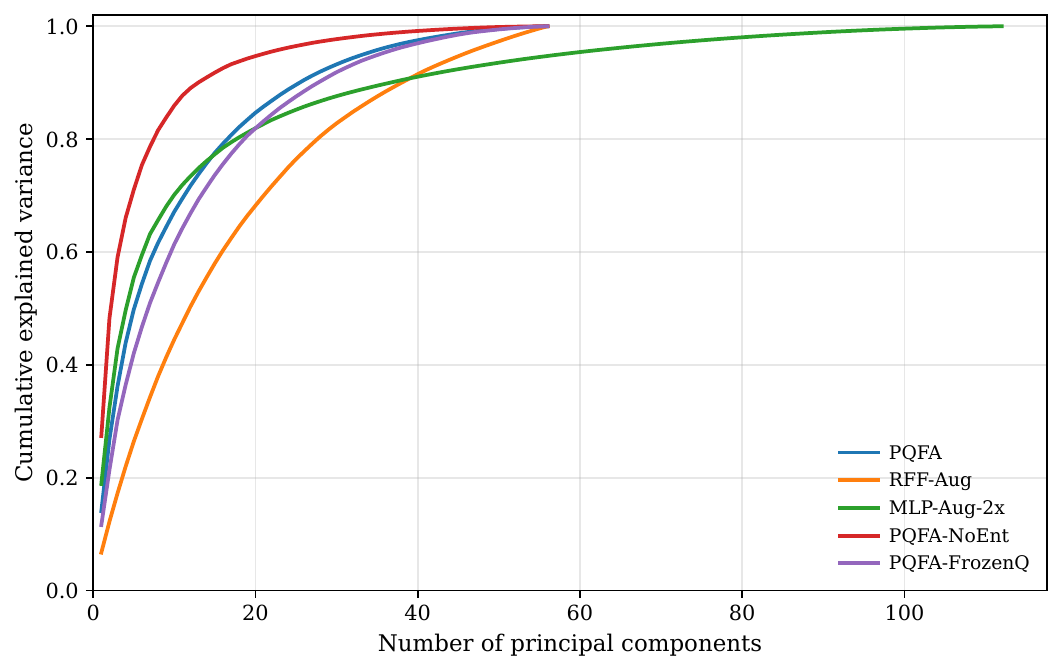}
\caption{PCA cumulative explained variance of augmentation features on MM-IMDb. Faster saturation indicates stronger variance concentration in fewer principal directions, whereas slower saturation indicates a more dispersed augmentation feature space.}
\label{fig:aug_pca_cumulative}
\end{figure}

\begin{table*}[t]
\centering
\small
\caption{PCA-based feature-space diagnostics of augmentation variants on MM-IMDb. Values are averaged across random seeds. $d_{\mathrm{eff}}$ denotes the PCA-based effective dimension, and $k_{90}$ and $k_{95}$ denote the numbers of principal components needed to explain 90\% and 95\% of the augmentation-feature variance, respectively.}
\label{tab:aug_feature_space}
\setlength{\tabcolsep}{6pt}
\begin{tabular}{lccccc}
\toprule
Variant & $d_{\mathrm{eff}}$ & $k_{90}$ & $k_{95}$ & Micro-F1 & Macro-F1 \\
\midrule
PQFA & 15.44 & 26 & 34 & 68.28 & 61.85 \\
RFF-Aug & 33.29 & 39 & 46 & 67.78 & 60.47 \\
MLP-Aug-2x & 12.56 & 37 & 58 & 67.98 & 60.95 \\
PQFA-NoEnt & 6.97 & 13 & 21 & 67.50 & 61.35 \\
PQFA-FrozenQ & 19.50 & 28 & 36 & 67.70 & 60.23 \\
\bottomrule
\end{tabular}
\end{table*}

Fig.~\ref{fig:aug_pca_cumulative} and Table~\ref{tab:aug_feature_space} show that PQFA produces a moderately compact augmentation feature space. It does not have the largest effective dimension. RFF-Aug yields a substantially more dispersed feature space, with $d_{\mathrm{eff}}=33.29$, but its test performance is lower than PQFA. This suggests that a larger feature-space spread alone is insufficient to explain the improvement of PQFA.

The no-entanglement quantum variant shows the strongest variance concentration, with $d_{\mathrm{eff}}=6.97$ and $k_{95}=21$. This indicates that removing entangling operations restricts the diversity of the generated quantum readouts, and this restricted augmentation space coincides with weaker test performance. By contrast, PQFA-FrozenQ has a comparable or even larger feature-space spread than PQFA, but its classification performance is lower. This suggests that random quantum transformations alone are insufficient; task-driven optimization of the quantum branch is important for aligning the readout features with the downstream objective.

The wider classical augmentation variant, MLP-Aug-2x, also provides a useful diagnostic. Although it increases the classical augmentation output dimension to 112 and contains 31,216 augmentation-branch parameters, it does not outperform PQFA. This observation further indicates that the improvement of PQFA is not simply a consequence of adding more augmentation dimensions or increasing the feature width.

Overall, these feature-space diagnostics support the interpretation that PQFA acts as a structured post-fusion feature transformation rather than a generic width-expansion module. The quantum branch produces an augmentation space that is neither excessively concentrated, as in PQFA-NoEnt, nor merely dispersed, as in RFF-Aug. Instead, it forms a moderately compact and task-effective nonlinear representation of the fused multimodal feature.

\subsection{Quantum-state Diagnostics}
\label{sec:quantum_state_diagnostics}

The preceding analyses evaluate whether the quantum augmentation branch improves prediction and whether its output features differ from generic width expansion. We further examine the trained quantum branch from a quantum-state perspective. Unlike the ablation experiments, the following diagnostics do not change the model architecture and do not involve additional training. We fix the trained PQFA model and conduct post-training noisy inference and noiseless entanglement entropy analysis. The purpose is to examine whether the learned quantum readout branch remains stable under simulated quantum perturbations and whether the parallel brickwall circuits induce nontrivial state-structure modulation of the amplitude-encoded fused representation.

\subsubsection{Post-training Noisy Inference}

We first evaluate the robustness of the trained quantum readout branch under simulated inference-time noise. All learned parameters are fixed. For each test sample, the fused representation is computed by the classical multimodal backbone and then passed through the trained quantum branches. During noisy inference, a single-qubit noise channel is inserted into the quantum branch after the local brickwall operations. This setting evaluates the sensitivity of the learned quantum readouts without performing noise-aware retraining.

Let $\rho$ denote a single-qubit density matrix. We consider five noise channels: depolarizing noise, amplitude damping, phase damping, bit-flip noise, and phase-flip noise. The Pauli operators are written as
\begin{equation}
\begin{aligned}
X &= |0\rangle\langle 1|+|1\rangle\langle 0|,\\
Y &= -i|0\rangle\langle 1|+i|1\rangle\langle 0|,\\
Z &= |0\rangle\langle 0|-|1\rangle\langle 1|.
\end{aligned}
\end{equation}

The bit-flip channel applies a Pauli-$X$ error with probability $p$:
\begin{equation}
\mathcal{E}_{\mathrm{bf},p}(\rho)
=
(1-p)\rho+pX\rho X .
\end{equation}
The phase-flip channel applies a Pauli-$Z$ error with probability $p$:
\begin{equation}
\mathcal{E}_{\mathrm{pf},p}(\rho)
=
(1-p)\rho+pZ\rho Z .
\end{equation}
The depolarizing channel replaces the clean state with Pauli-error states with total probability $p$:
\begin{equation}
\mathcal{E}_{\mathrm{dep},p}(\rho)
=
(1-p)\rho
+
\frac{p}{3}
\left(
X\rho X+Y\rho Y+Z\rho Z
\right).
\end{equation}

Amplitude damping models relaxation from $|1\rangle$ to $|0\rangle$. Its Kraus operators are
\begin{equation}
A_0
=
|0\rangle\langle 0|
+
\sqrt{1-p}|1\rangle\langle 1|,
\qquad
A_1
=
\sqrt{p}|0\rangle\langle 1|.
\end{equation}
The corresponding channel is
\begin{equation}
\mathcal{E}_{\mathrm{ad},p}(\rho)
=
A_0\rho A_0^{\dagger}
+
A_1\rho A_1^{\dagger}.
\end{equation}

Phase damping suppresses coherence while preserving computational-basis populations. Its Kraus operators are
\begin{equation}
P_0
=
|0\rangle\langle 0|
+
\sqrt{1-p}|1\rangle\langle 1|,
\qquad
P_1
=
\sqrt{p}|1\rangle\langle 1|.
\end{equation}
The corresponding channel is
\begin{equation}
\mathcal{E}_{\mathrm{pd},p}(\rho)
=
P_0\rho P_0^{\dagger}
+
P_1\rho P_1^{\dagger}.
\end{equation}

For an $n$-qubit circuit, these channels are applied as local quantum operations. If $\mathcal{E}_{\nu,p}$ is applied to the $r$-th qubit, the corresponding full operation is
\begin{equation}
\mathcal{E}_{\nu,p}^{(r)}
=
\mathcal{I}^{\otimes(r-1)}
\otimes
\mathcal{E}_{\nu,p}
\otimes
\mathcal{I}^{\otimes(n-r)},
\end{equation}
where $\nu\in\{\mathrm{dep},\mathrm{ad},\mathrm{pd},\mathrm{bf},\mathrm{pf}\}$. In our experiments, the selected channel is inserted into each trained quantum branch after the local brickwall operations. Since the noisy circuit generally produces a mixed state, the Pauli-$Z$ readout is computed as
\begin{equation}
q_{k,r}^{(\nu,p)}
=
\mathrm{Tr}
\left(
\rho_k^{(\nu,p)} O_r
\right),
\end{equation}
where $\rho_k^{(\nu,p)}$ is the noisy output state of the $k$-th quantum branch under noise channel $\nu$ and probability $p$, and $O_r$ is the single-qubit Pauli-$Z$ observable defined in Eq.~\eqref{eq:pauli_z_observables}. The noise probability is selected from
\begin{equation}
p \in \{0.001, 0.005, 0.01, 0.02, 0.05\}.
\end{equation}
The clean analytic inference result is used as the reference.

Let $q_i^{(0)}$ denote the clean Pauli-$Z$ readout vector of the $i$-th test sample, and let $q_i^{(\nu,p)}$ denote the corresponding readout under noise channel $\nu$ with probability $p$. We quantify the perturbation of the quantum branch by the average quantum readout drift:
\begin{equation}
D_q(\nu,p)
=
\frac{1}{N}
\sum_{i=1}^{N}
\left\|
q_i^{(\nu,p)}-q_i^{(0)}
\right\|_2 .
\end{equation}
We also report the probability drift and the label flip rate relative to clean analytic inference. The label flip rate measures the fraction of binary label decisions that change after injecting noise.

\begin{table*}[t]
\centering
\small
\caption{Post-training noisy inference results on MM-IMDb at $p=0.05$. All model parameters are fixed. $\Delta$Micro-F1 and $\Delta$Macro-F1 are measured relative to clean analytic inference and reported in percentage points.}
\label{tab:quantum_noise_summary}
\setlength{\tabcolsep}{6pt}
\begin{tabular}{lcccccc}
\toprule
Noise channel & Micro-F1 & Macro-F1 & $\Delta$Micro-F1 & $\Delta$Macro-F1 & $D_q$ & Flip rate \\
\midrule
Amplitude damping & 68.09 & 61.51 & -0.19 & -0.34 & 0.602 & 0.894\% \\
Bit flip & 68.16 & 61.75 & -0.12 & -0.10 & 0.591 & 0.605\% \\
Depolarizing & 68.11 & 61.69 & -0.17 & -0.16 & 0.595 & 0.528\% \\
Phase damping & 68.25 & 61.86 & -0.03 & +0.02 & 0.235 & 0.168\% \\
Phase flip & 68.15 & 61.66 & -0.14 & -0.18 & 0.571 & 0.591\% \\
\bottomrule
\end{tabular}
\end{table*}

\begin{figure*}[t]
\centering
\includegraphics[width=0.95\textwidth]{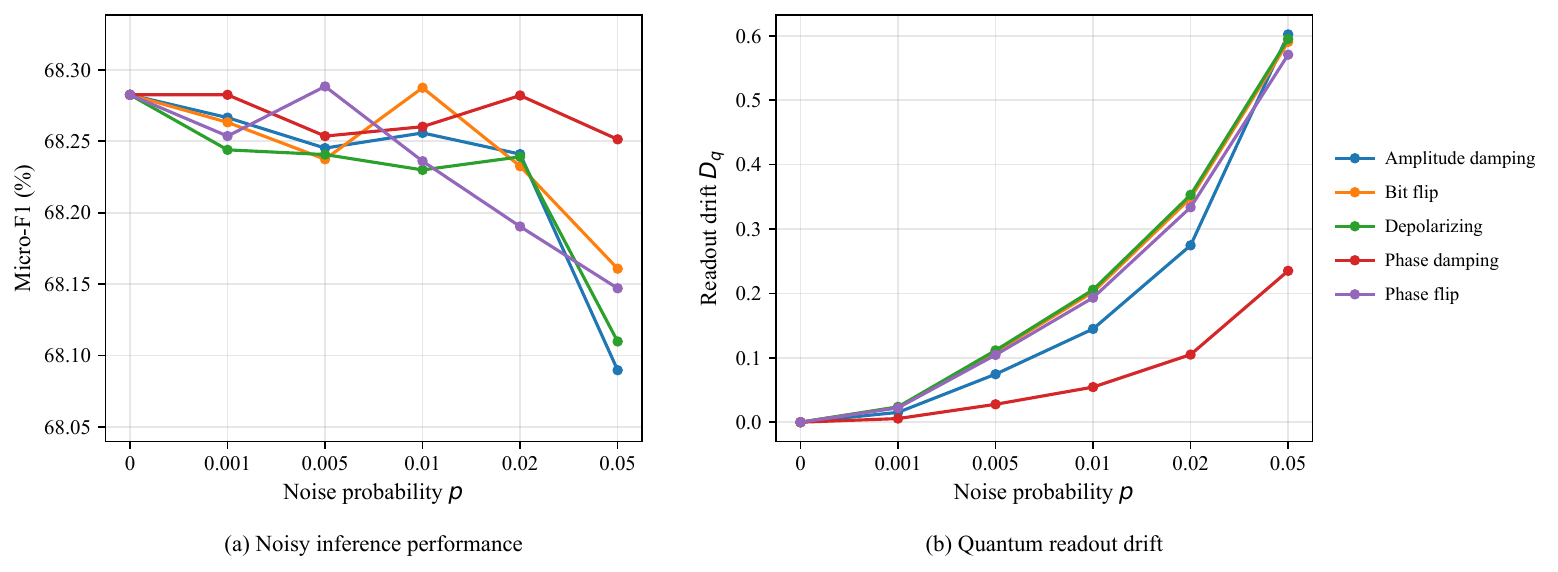}
\caption{Post-training noisy inference analysis on MM-IMDb. Panel (a) reports Micro-F1 under different noise probabilities, and panel (b) reports the quantum readout drift \(D_q\). All model parameters are fixed during this analysis.}
\label{fig:quantum_noise_diagnostics}
\end{figure*}

Table~\ref{tab:quantum_noise_summary} reports the noisy inference results at $p=0.05$, and Fig.~\ref{fig:quantum_noise_diagnostics} shows the full curves over different noise probabilities. The clean PQFA model obtains 68.28 Micro-F1 and 61.85 Macro-F1 on MM-IMDb. Under all five simulated noise channels, the final classification performance remains stable. Even at $p=0.05$, the Micro-F1 degradation is below 0.20 percentage points for all channel types. The full curves also show that the performance varies only within a narrow range around the clean result. Therefore, moderate inference-time perturbations of the quantum branch do not lead to a substantial collapse of the final multimodal classifier.

At the same time, the quantum readout drift $D_q$ increases clearly as the noise probability becomes larger. This trend is visible for all five noise channels, indicating that the injected noise channels produce nontrivial perturbations to the Pauli-$Z$ measurement features. The important point is that this readout perturbation does not translate into a comparable degradation of Micro-F1. This separation between readout-level perturbation and prediction-level stability indicates that the downstream classifier is tolerant to moderate quantum-readout fluctuations under the tested simulated noise channels.

The five channels exhibit different perturbation patterns. Phase damping has the weakest effect on the final prediction and also produces the smallest quantum readout drift. In contrast, amplitude damping, depolarizing noise, bit-flip noise, and phase-flip noise produce larger readout drift at high noise probabilities. Amplitude damping yields the largest Macro-F1 drop and the highest label flip rate at $p=0.05$, which is consistent with its direct effect on the population structure of the measured quantum state. Taken together, the noisy inference results show measurable quantum-readout sensitivity without a comparable collapse in prediction performance.

\subsubsection{Entanglement Entropy Analysis}

We further analyze the noiseless pure-state circuits using bipartite von Neumann entanglement entropy. For a quantum state $|\psi\rangle$ and a subsystem $A$, the reduced density matrix is
\begin{equation}
\rho_A=\mathrm{Tr}_{\bar{A}}\left(|\psi\rangle\langle\psi|\right),
\end{equation}
and the entanglement entropy is defined as
\begin{equation}
S(A)=-\mathrm{Tr}\left(\rho_A\log_2\rho_A\right).
\end{equation}
We use the main cut $A=\{0,1,2\}$ and $\bar{A}=\{3,4,5,6\}$. Entropy is computed after amplitude embedding and after each brickwall layer. Since this diagnostic aims to characterize the state structure induced by the learned quantum branches, it is performed on noiseless pure-state simulations rather than on mixed states under noise.

\begin{figure*}[t]
\centering
\includegraphics[width=0.95\textwidth]{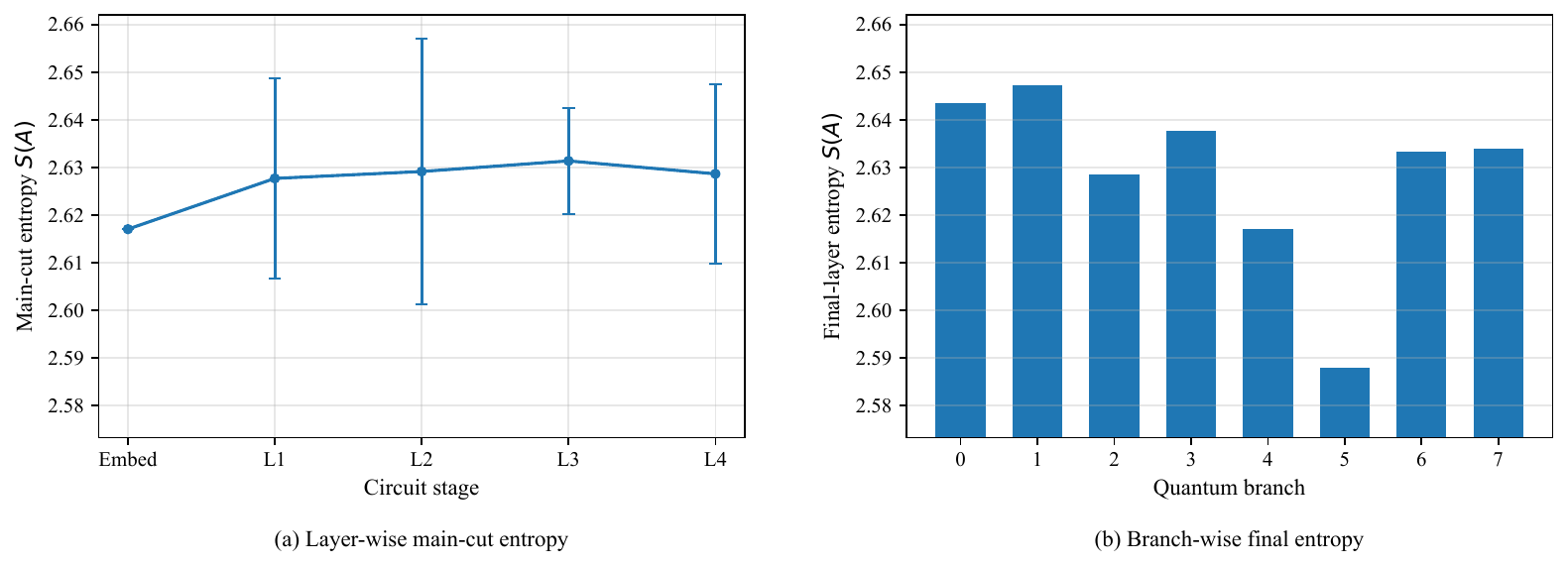}
\caption{Entanglement entropy diagnostics of the trained PQFA quantum branches. Panel (a) shows layer-wise main-cut entropy averaged across branches, and panel (b) shows final-layer main-cut entropy for each branch.}
\label{fig:entanglement_diagnostics}
\end{figure*}

Fig.~\ref{fig:entanglement_diagnostics} shows the layer-wise and branch-wise entropy results. Amplitude embedding already induces a highly entangled state, with an average main-cut entropy of 2.62. The trained brickwall layers further modulate this entropy moderately: the average main-cut entropy increases from 2.617 after amplitude embedding to 2.629 after the final layer. The single-qubit average entropy also increases from 0.975 to 0.981. Thus, the trained quantum layers should be interpreted as modulating an already entangled amplitude-encoded representation, rather than as the sole source of entanglement in the model.

The layer-wise curve shows a mild increase from the embedding stage to the intermediate brickwall layers, followed by a small decrease at the final layer. This indicates that the trained shallow circuits adjust the entanglement profile rather than monotonically maximizing it. The branch-wise result further shows that different parallel branches induce different final entanglement levels. Branches 0 and 1 yield relatively high final main-cut entropy, whereas branch 5 produces a lower final entropy than the other branches. This branch-dependent behavior indicates that the parallel brickwall circuits do not act as identical repeated modules. Rather, they provide structurally diverse quantum-state transformations of the same fused multimodal representation.

Together with the feature-space diagnostics in Section~\ref{sec:feature_space_diagnostics}, these results support the interpretation that PQFA acts as a structured post-fusion quantum augmentation module rather than a generic feature-width expansion. The noisy inference analysis shows stability of the learned readout under simulated perturbations, while the entanglement analysis shows moderate and heterogeneous state-structure modulation across the parallel quantum branches.

\subsection{Gate Weight Analysis}
\label{sec:gate_analysis}

\begin{figure*}[t]
  \centering
  \includegraphics[width=0.9\textwidth]{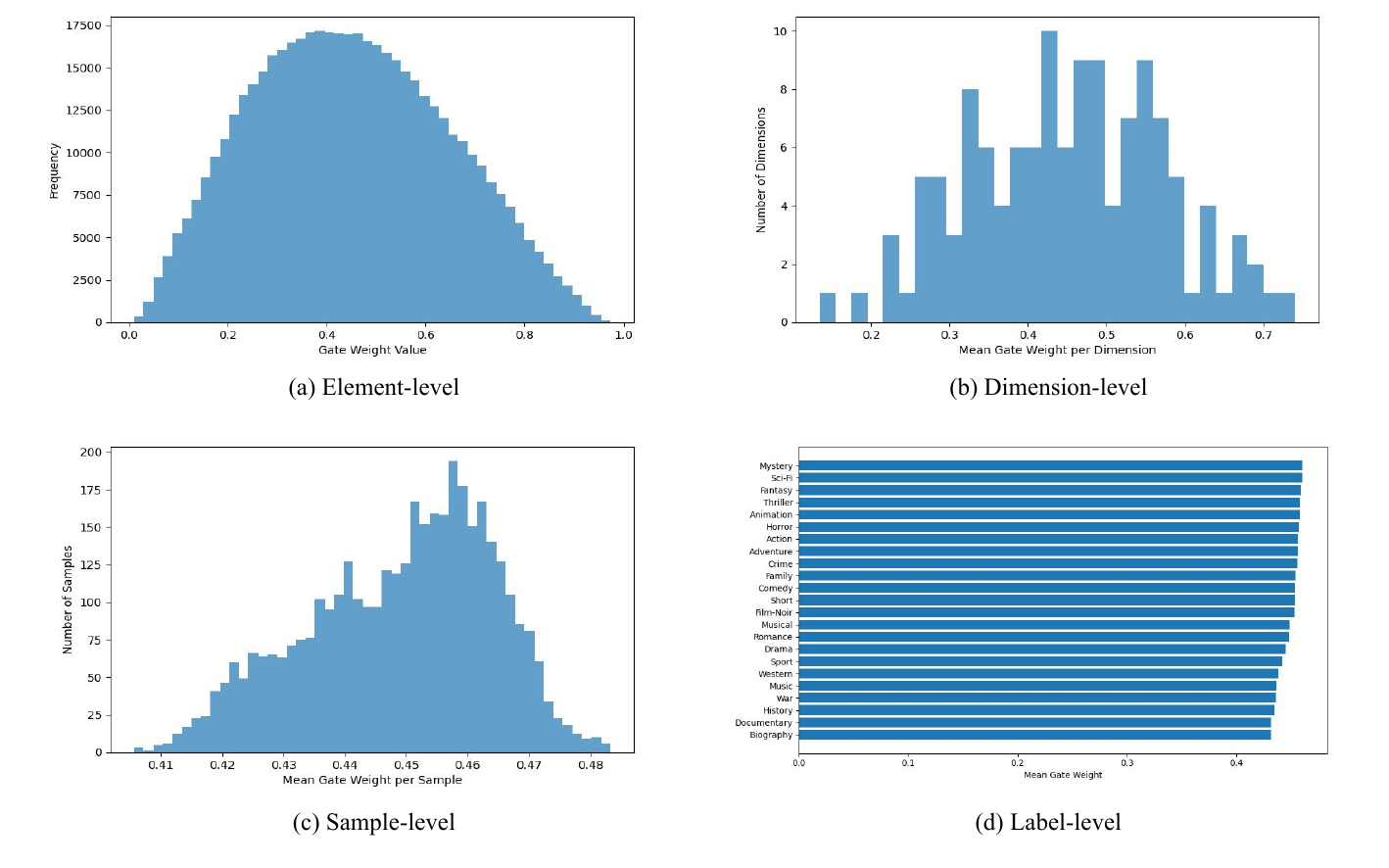}
  \caption{
  Gate weight distributions on MM-IMDb.
  (a) Element-level distribution over all samples and dimensions.
  (b) Dimension-level mean gate weights.
  (c) Sample-level mean gate weights.
  (d) Label-level mean gate weights across genre categories.
  }
  \label{fig:gate_analysis}
\end{figure*}

To further understand the behavior of the adaptive gated fusion module, we analyze the learned gate weights on the test set. 
Recall that the fused representation is computed as
\begin{equation}
\mathbf{z}_C = \mathbf{g}\odot \mathbf{z}_T + (1-\mathbf{g})\odot \mathbf{z}_I,
\end{equation}
where $\mathbf{g}$ denotes the learned gate vector. 
A larger gate value indicates a stronger contribution from the text-side representation $\mathbf{z}_T$, while a smaller gate value indicates a stronger contribution from the image-side representation $\mathbf{z}_I$. 
It should be emphasized that $\mathbf{z}_T$ and $\mathbf{z}_I$ are not raw unimodal features, but cross-modal representations obtained after bidirectional attention. 
Therefore, the gate values should be interpreted as the routing weights between two cross-modal representation streams rather than as direct causal attributions of raw text and image importance.

Fig.~\ref{fig:gate_analysis} presents the gate behavior at four different granularities. 
At the element level, Fig.~\ref{fig:gate_analysis}(a) shows that the gate values span a broad range rather than concentrating around a single constant value. 
This indicates that the fusion module does not degenerate into a fixed averaging strategy or a hard selection of one modality. 
Meanwhile, the distribution is centered below $0.5$, suggesting that the fused representation tends to assign relatively larger weights to the image-side representation under complete multimodal inputs.

The dimension-level analysis in Fig.~\ref{fig:gate_analysis}(b) further reveals heterogeneous modality preferences across latent dimensions. 
Although many dimensions show an image-side preference, a non-negligible subset of dimensions has mean gate values above $0.5$, indicating stronger reliance on the text-side representation. 
This suggests that the adaptive gate performs feature-wise routing: different latent dimensions specialize in different modality streams instead of applying a uniform global modality weight.

At the sample level, Fig.~\ref{fig:gate_analysis}(c) shows that the mean gate values are more concentrated and remain mostly below $0.5$. 
This implies that, for most test samples, the overall fused representation is mildly biased toward the image-side stream. 
However, compared with the element-level and dimension-level distributions, the sample-level distribution is much narrower, suggesting that the main variation of the gate occurs at the feature-dimension level rather than as large sample-wise shifts between text-dominant and image-dominant fusion.

The label-level results in Fig.~\ref{fig:gate_analysis}(d) show a similar overall tendency. 
Across genres, the mean gate values are generally below $0.5$, indicating a consistent image-side preference in the fused representation. 
Nevertheless, different genres exhibit small but observable variations. 
Genres with relatively larger gate values rely more on the text-side stream, whereas genres with smaller gate values rely more on the image-side stream. 
This suggests that the learned fusion behavior is not completely category-agnostic, but reflects certain category-dependent differences in modality usage.

This gate analysis is complementary to the missing-modality experiments. 
The missing-modality results evaluate the functional dependence of the whole model under input-level perturbations, whereas the gate analysis describes how the fusion module routes information between two cross-modal representation streams when both modalities are available. 
Therefore, a stronger performance degradation caused by removing text does not necessarily imply that the gate weights should be globally larger than $0.5$. 
Textual information can still be crucial for prediction because it contributes to the cross-modal representations before gated fusion and because several latent dimensions remain text-oriented. 
At the same time, the gate statistics show that, once both modalities have been encoded and interacted through cross-attention, the fusion module tends to assign larger weights to the image-side representation. 
This indicates that the model does not simply rely on a single dominant modality, but combines textual semantics and visual evidence through structured feature-wise routing.

These observations provide a diagnostic view of the classical fusion behavior before quantum feature augmentation.
They indicate that the input to the quantum branch is not a raw unimodal representation, but a fused feature shaped by cross-modal interaction and adaptive feature-wise routing.

\section{Discussion and Future Work}
\label{sec:discussion}

The main finding of this study is that shallow parallel quantum circuits can serve as parameter-efficient post-fusion augmentation modules for multimodal classification. The primary evidence comes from the controlled comparisons with NoQ and MLP-Aug, in which the pretrained encoders, data splits, projection dimension, fusion backbone, and training protocol for the shared classical components are held fixed. Under these conditions, the improvement achieved by PQFA cannot be explained by stronger encoders, a different fusion mechanism, a simple increase in the final representation width, or a larger parameter budget in the augmentation branch. Instead, the results indicate that the measurement readouts generated by the parallel quantum branches provide a useful nonlinear transformation of the fused multimodal representation.

From the perspective of multimodal fusion, PQFA addresses a stage of the standard pipeline that has received comparatively limited attention. Even after cross-modal attention and adaptive gating have produced an informative fused representation, a compact post-fusion transformation can extract additional discriminative features that are not fully captured by applying the final classifier directly to the fused feature.

The controlled comparisons on N24News provide further support for this interpretation. The width-matched MLP-Aug baseline slightly underperforms the NoQ backbone, achieving 84.23\% accuracy compared with 84.70\%. This result shows that simply concatenating additional learned nonlinear features with the fused representation does not necessarily improve prediction and may introduce features that are not well aligned with the task. By contrast, PQFA reaches 85.35\% accuracy and improves upon the same NoQ backbone. This difference indicates that the benefit of PQFA does not arise from feature expansion alone, but from the learned transformation provided by the quantum readout branches. In addition to its predictive advantage, the PQFA augmentation module requires approximately 2.2K trainable parameters, compared with 24.0K for MLP-Aug, demonstrating that the improvement is achieved with a substantially smaller augmentation-branch parameter budget.

The missing-modality results extend this finding from clean-input performance to robustness under incomplete observations. Although PQFA already improves performance when both modalities are available, its advantage becomes particularly clear when the more informative textual modality is severely degraded. This indicates that the quantum augmentation branch contributes not only to complete-input prediction, but also to maintaining useful decision features when part of the multimodal evidence is unavailable. The paired error-transition analysis provides a complementary decision-level perspective. Relative to MLP-Aug, PQFA produces a stronger positive correction pattern, while its changes relative to the already competitive NoQ backbone remain predominantly beneficial. These findings position PQFA as a complementary post-fusion feature transformation that strengthens, rather than replaces, the classical fusion backbone.

The feature-space diagnostics further clarify the nature of the augmentation effect. PQFA does not generate the most widely dispersed feature space: RFF-Aug exhibits a larger effective dimension, yet achieves lower test performance. A larger feature-space spread is therefore not sufficient to explain the gains of PQFA. In contrast, PQFA-NoEnt produces a strongly concentrated augmentation space together with weaker predictive performance, suggesting that removing the trainable entangling operations restricts the diversity and utility of the measurement features. PQFA-FrozenQ produces a feature-space spread comparable to that of the trained model but performs less effectively, showing that random quantum transformations alone cannot reproduce the benefit of task-driven optimization. Taken together, these results indicate that PQFA is not simply a mechanism for increasing dimensionality. Rather, it produces a compact, structured, and task-aligned nonlinear transformation of the fused representation.

The missing-modality and gate-weight analyses capture two distinct aspects of modality use. The missing-modality experiments measure functional dependence by perturbing one input modality during inference, whereas the gate-weight analysis examines how the fusion module routes information between the two cross-modal representation streams when both modalities are present. Consequently, the larger performance decline caused by removing text does not require the learned gate values to be globally dominated by the textual stream. Textual information can influence both representation streams through bidirectional cross-attention before gated fusion, and individual latent dimensions may retain different degrees of text- or image-related information. The complete-input gate statistics may therefore assign relatively larger weights to the image-side stream even when the final prediction remains more sensitive to the removal of text. Together, these analyses show that the quantum branches operate on an integrated representation shaped by both cross-modal interaction and feature-wise routing, rather than on a raw unimodal feature.

The current study also identifies several directions for further investigation. The post-training noise experiments evaluate the sensitivity of the learned quantum readouts to simulated channel perturbations, but all results are obtained through classical simulation rather than execution on physical quantum hardware. Such simulations do not fully capture device-specific connectivity, calibration drift, readout errors, compilation overhead, finite-shot measurement variability, or the practical cost of amplitude state preparation. Future work can therefore extend PQFA to hardware-aware settings that incorporate realistic circuit compilation, finite-shot measurements, device-calibrated noise models, and readout-error mitigation. Comparisons with classical dropout and noise-injection baselines would also help distinguish hardware-induced robustness from robustness arising from the structure of the augmentation module itself. These extensions would provide a more complete assessment of the practical behavior of post-fusion quantum augmentation on near-term quantum platforms.

\section{Conclusion}
\label{sec:conclusion}

This paper introduced Parallel Quantum Feature Augmentation (PQFA), a post-fusion framework that uses shallow parallel variational quantum circuits to augment fused multimodal representations. PQFA preserves the classical encoding and fusion backbone while incorporating quantum measurement readouts as additional features for prediction. Experiments on MM-IMDb and N24News show that PQFA improves both multi-label and single-label classification over the no-quantum backbone and a width-matched MLP augmentation baseline. These gains are achieved with approximately 2.2K augmentation parameters, compared with 24.0K for MLP-Aug.
Controlled ablations and feature-space analyses indicate that the improvements cannot be explained by feature expansion, random transformations, or additional classical capacity alone. Missing-modality and paired-transition results further show that PQFA improves robustness under incomplete inputs and produces beneficial decision-level corrections. Simulated-noise and gate-weight analyses provide additional evidence on the stability of the quantum readouts and the multimodal structure of their inputs. Overall, the results demonstrate that shallow quantum circuits can serve as effective and parameter-efficient post-fusion augmentation modules without replacing strong classical fusion components. Future work will investigate hardware-aware implementations, finite-shot evaluation, circuit and branch scaling, and broader multimodal tasks.

\section*{\textbf{Funding}}
This work was supported by the National Natural Science Foundation of China (Grant Nos. 12341103, 62372444) and the National Key R\&D Program of China (Grant No. 2023YFA1009403).

\bibliographystyle{elsarticle-num}
\bibliography{bibliography}
\end{document}